%% file: main.tex
\documentclass[10pt,twocolumn,letterpaper]{article}

\usepackage[pagenumbers]{iccv}      

\usepackage[accsupp]{axessibility}

\input{preamble}

\definecolor{cvprblue}{rgb}{0.21,0.49,0.74}
\usepackage[pagebackref,breaklinks,colorlinks,allcolors=cvprblue]{hyperref}

\def\paperID{11177} 
\def\confName{ICCV}
\def\confYear{2025}

\title{Geometry Distributions}

\author{Biao Zhang\\
KAUST\\
{\tt\small biao.zhang@kaust.edu.sa}
\and
Jing Ren\\
ETH Zurich\\
{\tt\small jing.ren@inf.ethz.ch}
\and
Peter Wonka\\
KAUST\\
{\tt\small pwonka@gmail.com}
}

\begin{document}
\twocolumn[{
\renewcommand\twocolumn[1][]{#1}%
\maketitle\vspace{-1cm}
\begin{center}
\captionsetup{type=figure}
\setlength{\abovecaptionskip}{-0.1cm}
\begin{overpic}[trim=0cm 0cm 0cm 0cm,clip,width=1\linewidth,grid=false]{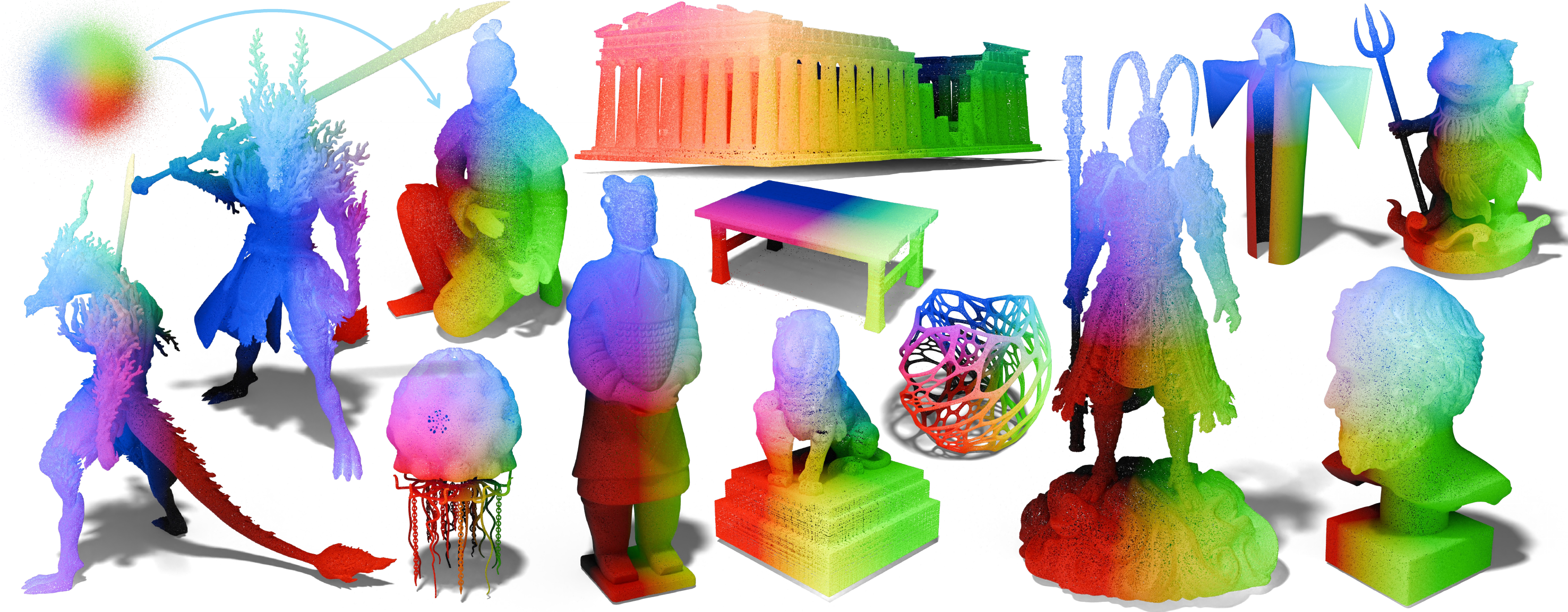}
\put(3,28){\scriptsize $\vec{X}\sim \mathcal{N}$}
\put(22,38){\scriptsize $\mathcal{E}_1(\vec{X})$}
\put(11,35.5){\scriptsize $\mathcal{E}_2(\vec{X})$}
\end{overpic}
\vspace{-7pt}
\captionof{figure}{Our representation can handle 3D geometry with complex details, high genus, sharp features, and non-watertight surfaces: our trained diffusion networks $\mathcal{E}_i$ can transform the samples $\vec{X}$ from a Gaussian distribution $\mathcal{N}$ to the geometry $\gM\subset\sR^3$. The colors indicate the correspondence between the Gaussian noise and the surface points.}
\label{fig:teaser}
\end{center}
}]


\begin{abstract}
Neural representations of 3D data have been widely adopted across various applications, particularly in recent work leveraging coordinate-based networks to model scalar or vector fields. 
However, these approaches face inherent challenges, such as handling thin structures and non-watertight geometries, which limit their flexibility and accuracy. 
In contrast, we propose a novel geometric data representation that models geometry as distributions---a powerful representation that makes no assumptions about surface genus, connectivity, or boundary conditions.
Our approach uses diffusion models with a novel network architecture to learn surface point distributions, capturing fine-grained geometric details. 
We evaluate our representation qualitatively and quantitatively across various object types, demonstrating its effectiveness in achieving high geometric fidelity. 
Additionally, we explore applications using our representation, such as textured mesh representation, neural surface compression, dynamic object modeling, and rendering, highlighting its potential to advance 3D geometric learning. 
\end{abstract}

\section{Introduction}
Geometry representations are at the heart of most 3D vision problems.
With the rapid advancement of deep learning, there is growing interest in developing neural network-friendly geometric data representations.
Recent advances in this field, particularly those based on coordinate networks, have shown promise in modeling 3D geometry for various applications, as their functional nature integrates well with neural networks.
However, they also face challenges like limited accuracy in capturing complex geometric structures and difficulties in handling non-watertight objects.

\begin{figure}[t]
    \centering
    \begin{overpic}[trim=0cm 0cm 0cm -0.3cm,clip,width=1\linewidth,grid=false]{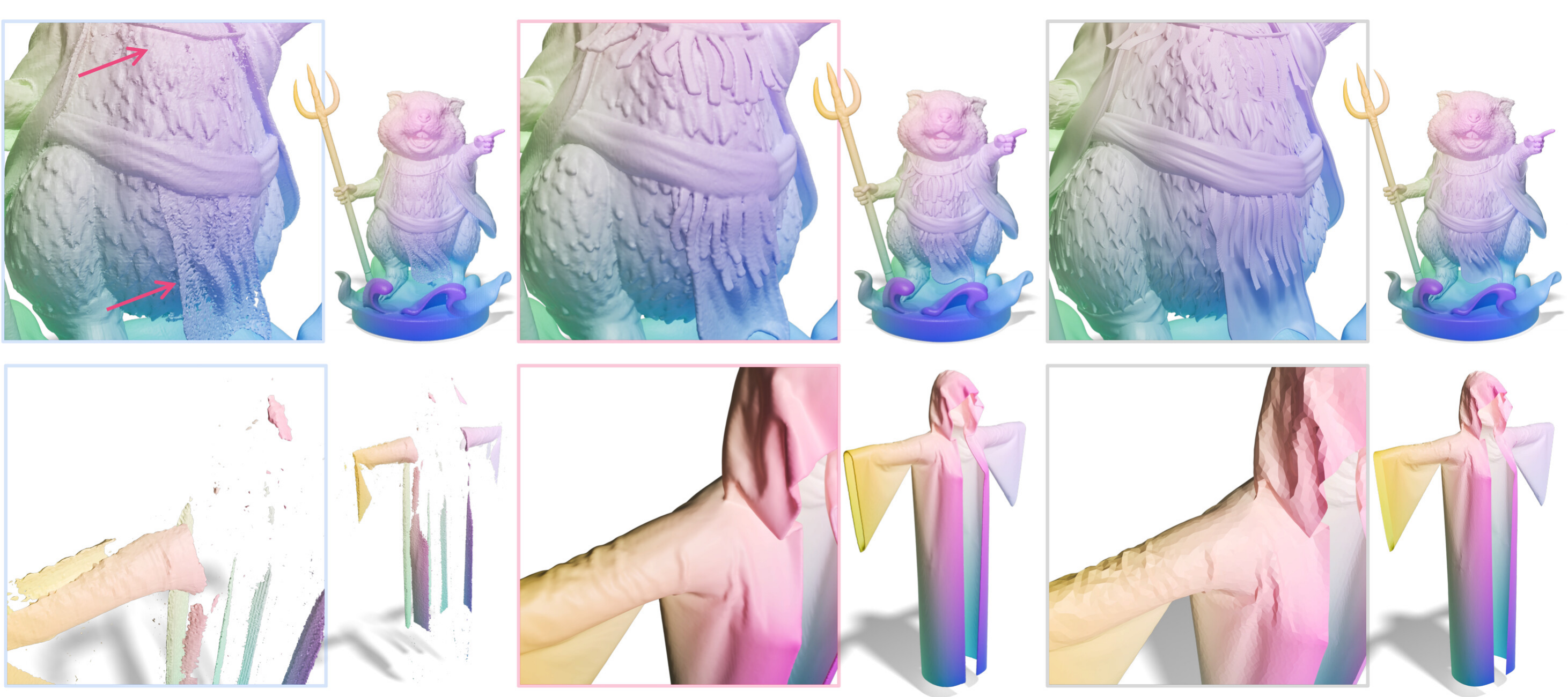}
    \put(12,48){\scriptsize SDF}
    \put(6, 45){\scriptsize{14M parameters}}
    \put(37,48){\scriptsize \coolnameshort (\textbf{ours})}
    \put(39, 45){\scriptsize{5M parameters}}
    \put(78,47){\scriptsize target surface}
    \end{overpic}
    \vspace{-16pt}
    \caption{Compared to signed distance functions (SDFs), our \coolnameshort can model open and non-watertight objects using significantly \emph{fewer} network parameters. See the Appendix for the meshing algorithm used in this figure. The SDF is fit using Instant-NGP~\cite{muller2022instant}, with isosurface extraction evaluated on a 512-resolution grid. We observe that SDFs struggle to represent thin structures or non-watertight geometry. }
    \label{fig:res:sdf}
\end{figure}

To overcome these challenges, we propose a new geometric data representation, possessing a simple and consistent data structure capable of accommodating shapes with varying genus, boundary conditions, and connectivity—whether open, watertight, fully connected, or not (see~\cref{fig:res:sdf}). 
A key insight is that any surface, regardless of its topology or structural integrity, can be closely approximated by a sufficiently large number of points sampled on the surface. 
Recent advancements in generative models have shown that, in theory, they can sample an infinite amount of data from a distribution. 
Building on these insights, we model 3D geometry as a \emph{distribution} of surface points, encoded into a diffusion model.
Unlike triangle mesh representations, which are specific discretizations of the underlying surface, or point clouds, which represent a particular sampling choice, our approach models the distribution of all possible surface points, providing a more continuous and accurate encoding of the underlying geometry.

Diffusion models, widely recognized for their effectiveness in 2D content generation, have emerged as a leading approach among generative models. However, their application to 3D geometry remains largely unexplored.
We found that 3D geometry context presents unique challenges: direct adaptation often falls short in capturing geometric details and results in inaccurate geometry recovery.

\begin{figure}[t]
    \centering
    \begin{overpic}[trim=0cm 0cm 0cm 0cm,clip,width=1\linewidth,grid=false]{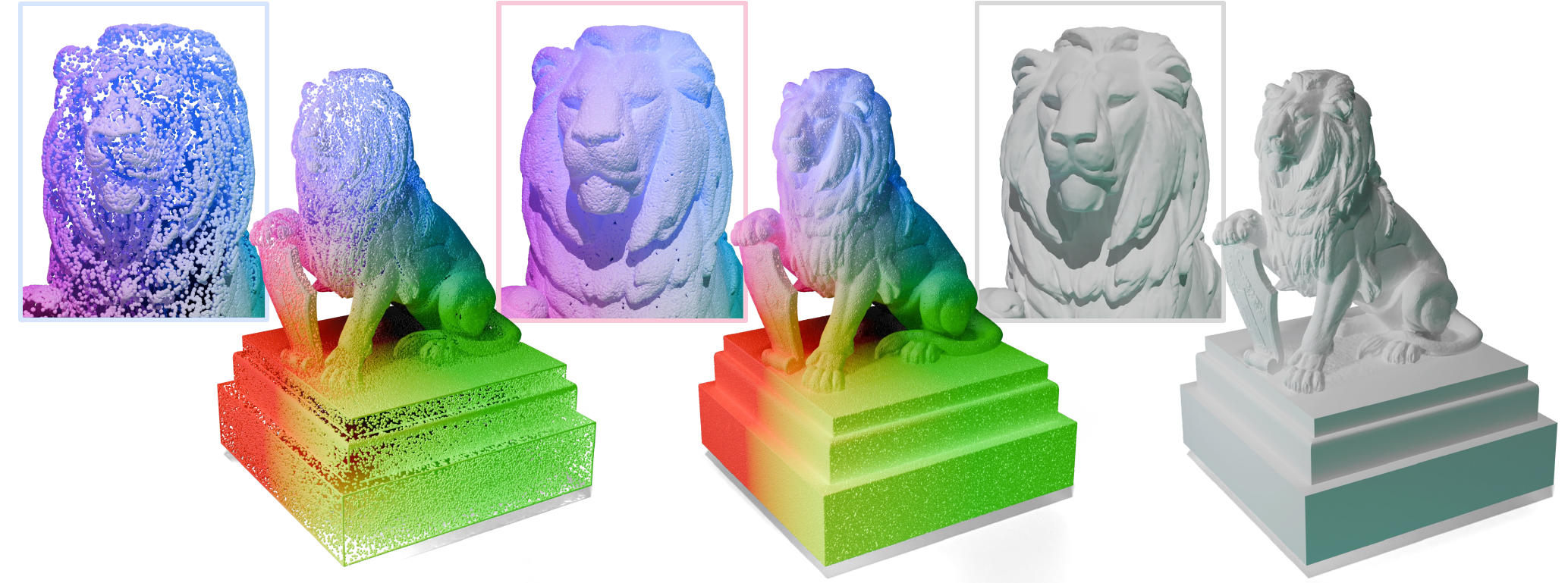}
    \put(10,39){\scriptsize vector fields}
    \put(34,39){\scriptsize \coolnameshort (\textbf{ours})}
    \put(72,39){\scriptsize target surface}
    \put(15, -2){\scriptsize CD = 4.886}
    \put(47, -2){\scriptsize CD = \textbf{3.218}}
    \end{overpic}
    \vspace{-10pt}
    \caption{
    Compared to vector fields-based method, our \coolnameshort produces more uniformly distributed samples with higher fidelity. The chamfer distance $(\times 10^3)$ between the samples and target surface is reported below. 
    }
    \label{fig:res:lion-vf-ours}
\end{figure}

In this work, we introduce \coolname (or \coolnameshort in short), a new representation for general geometric data. 
Our approach leverages a diffusion model with a novel network architecture.
\setlength{\columnsep}{5pt}%
\setlength{\intextsep}{1pt}%
\begin{wrapfigure}{r}{0.42\linewidth}
\centering
\begin{overpic}[trim=0cm 0.2cm 0cm 0cm,clip,width=1\linewidth,grid=false]{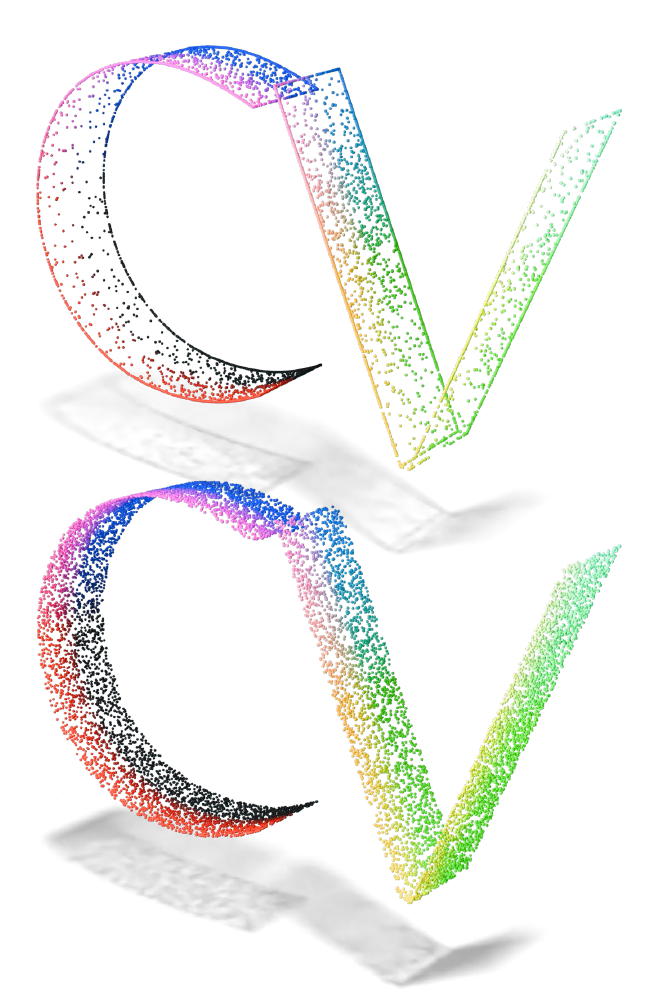}
\put(11,55){\scriptsize{vector fields}}
\put(11,11){\scriptsize{\coolnameshort}}
\end{overpic}\vspace{-3pt}
\end{wrapfigure}
By solving a forward ordinary differential equation (ODE), we map spatial points, sampled from Gaussian \emph{noise space}, to surface points in \emph{shape space}, enabling an infinite set of points to represent the geometry. This allows us to sample on the surface uniformly comparing to existing vector fields-based formulation (see the inset and \cref{fig:res:lion-vf-ours}).
Additionally, we derive the backward ODE algorithm, allowing for inverse mapping from the shape space back to noise space.
Our results demonstrate the accuracy and robustness of {\coolnameshort} across a broad range of complex structures. 
Furthermore, our approach enables the simultaneous encoding of texture or motion information alongside geometry.

To summarize, \coolnameshort facilitates a highly accurate yet compact neural representation of 3D geometry, demonstrating significant potential for future applications, including textured mesh representation, neural surface compression, dynamic object neural modeling, and photo-realistic rendering with Gaussian splatting. 

\begin{table}[b]
    \input{tables/tab_data_representation}
    \vspace{-4pt}
    \caption{Different representations for 3D geometric data.}\label{tab:data-representation}
\end{table}

\section{Related Works}
\subsection{Different representations for 3D geometry}\label{sec:rel:rep}
Existing geometry representations—such as meshes, voxels, point clouds, and implicit functions—each offer distinct advantages but also have inherent limitations. 
Triangle or polygonal meshes, which are commonly used in traditional geometry processing~\cite{botsch2010polygon}, are not ideal for geometric learning due to their inconsistent data structures when dealing with shapes that have a different number of vertices and different connectivity~\cite{bronstein2017geometric,fey2018splinecnn}. 

Voxels, with their inherent grid-based structure, are ideal for learning-based tasks. However, they are memory-intensive, especially when high resolution is needed for capturing fine details~\cite{liu2020neural,sun2022direct}. 
Point clouds, easily obtained from sensors, are widely used in geometric learning tasks~\cite{xiao2023unsupervised,guo2020deep,bello2020deep}. 
However, they are essentially samples of the geometry, leading to potential information loss of the underlying geometry.
Their expressiveness heavily depends on sampling density and uniformity, and the lack of inherent point connectivity complicates defining surface structures, boundaries, or geodesics along surfaces. 
Implicit functions~\cite{mescheder2019occupancy, park2019deepsdf} excel at generating smooth surfaces and representing complex topologies. However, they struggle with accurately modeling thin structures or non-watertight geometries (see~\cref{fig:res:sdf} for an illustration). 
Additionally, integrating colors or textures with implicit functions is not straightforward.

Our goal is to design a new data representation for various 3D geometric learning tasks, featuring a network-friendly data structure that accommodates shapes with varying genus, boundary conditions, and connectivity, whether open, watertight, fully connected, or not.  See~\cref{tab:data-representation} for a summary of different representations.

\subsection{Diffusion models}\label{sec:rel:diffusion}
Diffusion models are powerful generative models that transform data into noise through a forward diffusion process and learn to reverse this process to generate high-quality samples. 
Beginning with Denoising Diffusion Probabilistic Models~\cite{ho2020denoising}, diffusion models have evolved into more efficient and flexible approaches~\cite{song2020denoising, karras2022elucidating, liu2022flow, lipman2022flow}.
While our work does not contribute directly to advancements in diffusion models, we employ them as a foundation for modeling complex geometry. Our approach primarily builds upon the framework established in EDM~\cite{karras2022elucidating}.

Significant progress has been made in generating 3D geometry using diffusion models~\cite{hui2022neural, chen2022neural, shue20233d, zheng2023locally, vecset, ren2024xcube, yariv2024mosaic, xiong2024octfusion, dong2024gpld3d, petrov2024gem3d, zhang2024functional}, with most approaches representing geometry via signed distance functions or occupancy fields. 
Fewer methods, however, focus on point clouds~\cite{luo2021diffusion, zhou20213d, zeng2022lion} or Gaussian point clouds~\cite{roessle2024l3dg, zhang2024gaussiancube,yushigaussiananything}. 
We would also like to emphasize two related works~\cite{yang2019pointflow, cai2020learning}. While both works are capable of encoding infinite-resolution point clouds using normalizing flow~\cite{rezende2015variational} and score-based generative models~\cite{song2019generative}, they are only evaluated on toy datasets and do not demonstrate high-quality geometric details as we do.
These models are trained on a dataset of 3D objects, treating each \emph{object} as a single training sample. In contrast, our approach is fundamentally different, as we treat each \emph{spatial point} as an individual training sample.

\subsection{Coordinate-based neural representations}\label{sec:rel:coord}
Signed-distance functions (SDFs) are widely used to represent 3D geometry~\cite{takikawa2021neural, martel2021acorn, muller2022instant, sitzmann2020implicit, davies2020effectiveness, guan2023learning}. 
Instead of explicitly storing vertices or points, a network is trained to produce signed distances to the surface or signals indicating inside/outside for each spatial point, implicitly defining the shape's geometry. 
Although relatively easy to learn via neural networks, SDFs struggle to model non-watertight meshes.
Follow-up works are then introduced to model open surfaces, where the outputs of the networks are unsigned distances~\cite{chibane2020neural, guillard2022meshudf, yang2023neural, zhou2024cap} or vectors~\cite{yang2023neural, edavamadathil2024neural, yifan2021geometry} pointing toward the surface.
These works primarily use networks to fit scalar fields or vector fields, representing 3D data through networks that map coordinates to scalar or vector values. Our approach, while distinct in methodology, shares a conceptual connection with these works: it can be interpreted as a trajectory field. 



\subsection{Point-based graphics}\label{sec:rel:point}
Point-based computer graphics is an approach that represents 3D surfaces as sets of discrete points rather than traditional polygonal meshes. Unlike polygonal models that use vertices and edges to define shapes, point-based methods use individual points sampled across a surface to capture details directly. The field can be dated back to 1980s~\cite{levoy1985use}. Early works~\cite{pfister2000surfels, zwicker2001surface} investigated how to render with points. Until recently, works utilized point representations in differentiable rendering~\cite{yifan2019differentiable, kerbl20233d}. Different from our method, these works focus on rendering with finite number of points.

\begin{figure}[t]
    \centering
    \begin{overpic}[trim=0cm 0cm 0cm 0cm,clip,width=1\linewidth,grid=false]{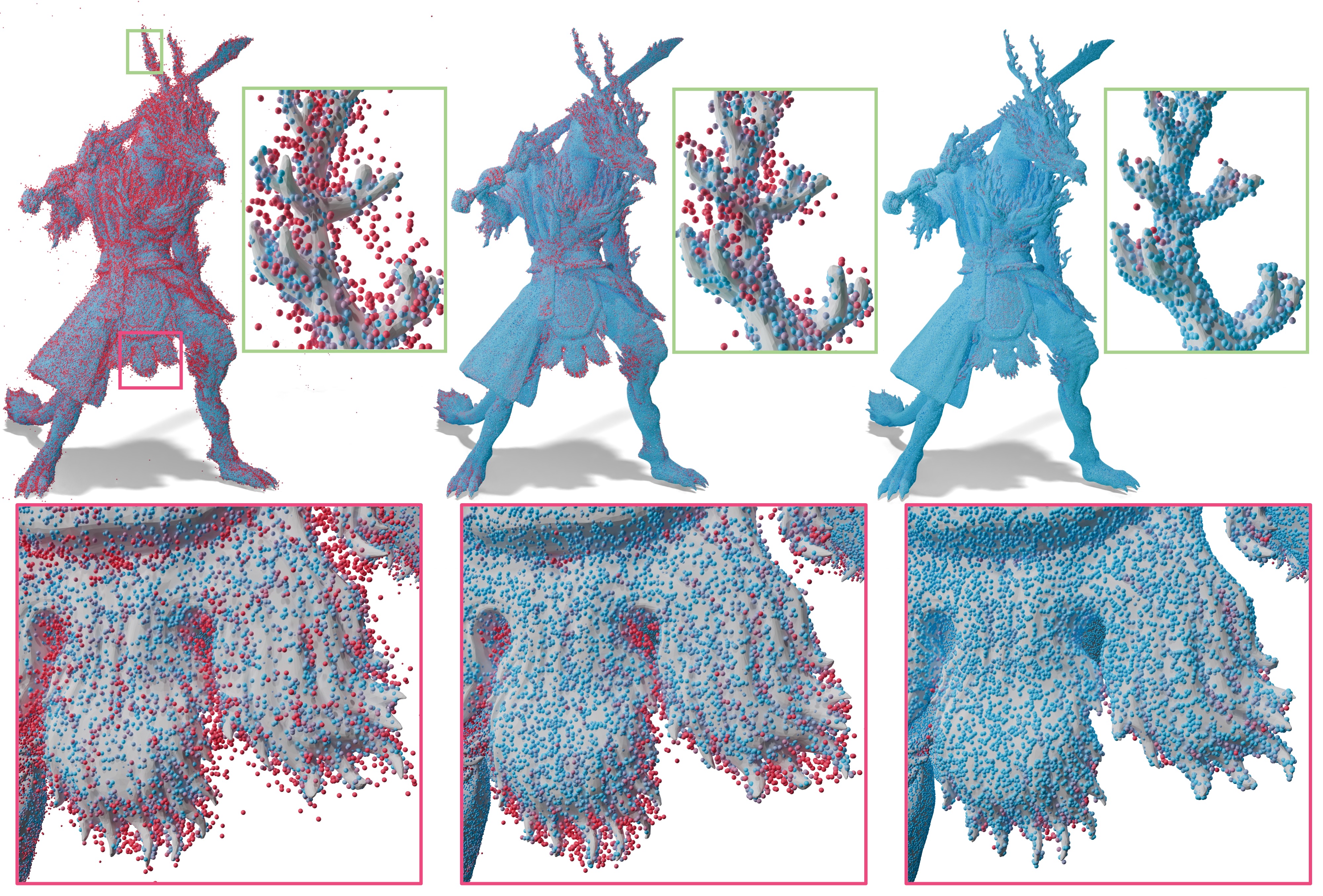}
    \put(10,67){\scriptsize hashing grids}
    \put(45,67){\scriptsize MLP}
    \put(80,67){\scriptsize \textbf{ours}}
    \end{overpic}\vspace{-6pt}
    \caption{Heatmap of the $L_2$ distance from sampled points to the target surface using different network architectures.}
    \label{fig:res:ours-vs-mlp}
\end{figure}

\section{Geometry Distributions}
\subsection{Problem formulation \& motivations}
Given a surface $\gM\subset\sR^3$, our goal is to model it as a probability distribution $\Phi_{\gM}$, such that any sample $\vec{x}\sim\Phi_{\gM}$ drawn from this distribution is a surface point, i.e., $\vec{x}\in\gM$. 
In this way, the distribution $\Phi_{\gM}$, which encodes the geometry $\gM$, provides a flexible geometric representation-any sampling, whether dense or sparse, closely approximates the surface $\gM$ at the target resolution. 
Inspired by the pioneering work ``Geometry Images'', which uses 2D images to represent 3D meshes~\cite{gu2002geometry}, we name our representation as \coolname.

\begin{figure*}[t]\vspace{-6pt}
    \includegraphics[width=1\linewidth]{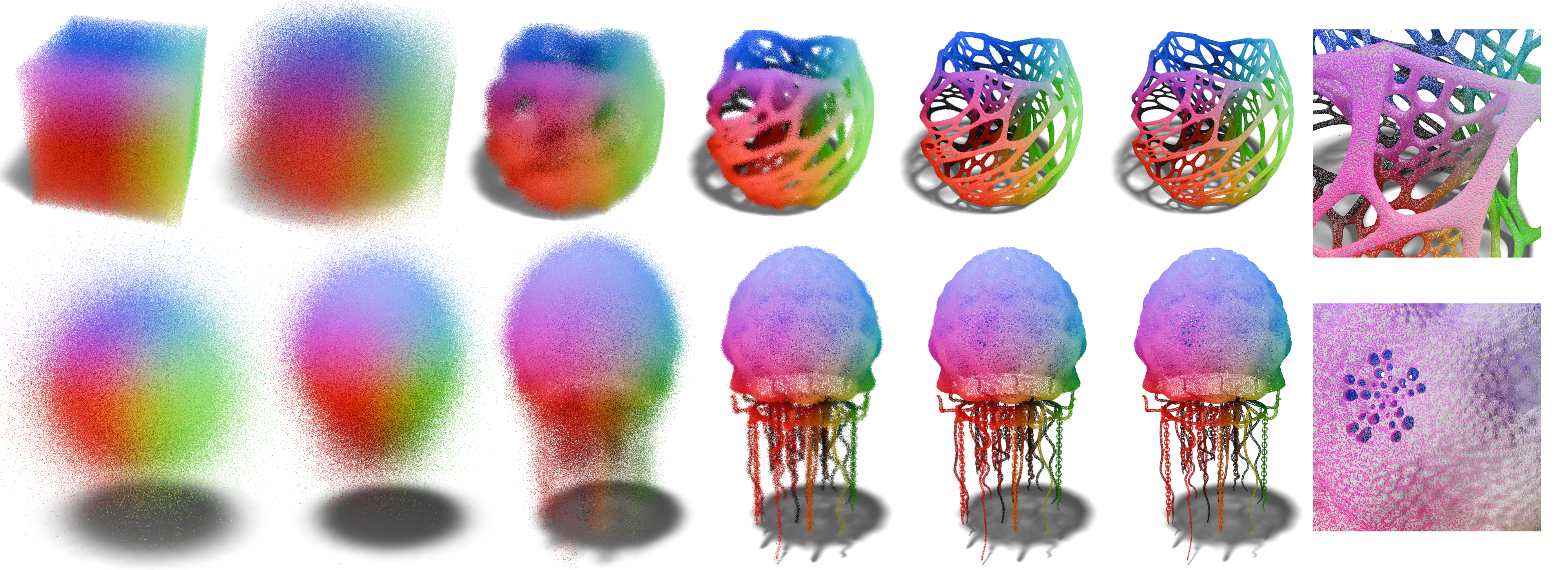}\vspace{-9pt}
    \caption{\textbf{Inference process} for generating 1M points on a lamp mesh (\emph{top}) and a jellyfish mesh (\emph{bottom}) from uniform and Gaussian distributions, respectively. Results are shown at timesteps $t = 0, 40, 48, 56, 60, 64$, with a close-up of the generated samples at $t=64$ overlaid on the ground-truth mesh. A complete illustration is available in the accompanying video demo. Both meshes are taken from~\cite{zhou2016thingi10k}.}
    \label{fig:res:lamp-jellyfish}
\end{figure*}

\begin{figure*}[t]
    \begin{overpic}[trim=0cm 0cm 0cm 0cm,clip,width=1\linewidth,grid=false]{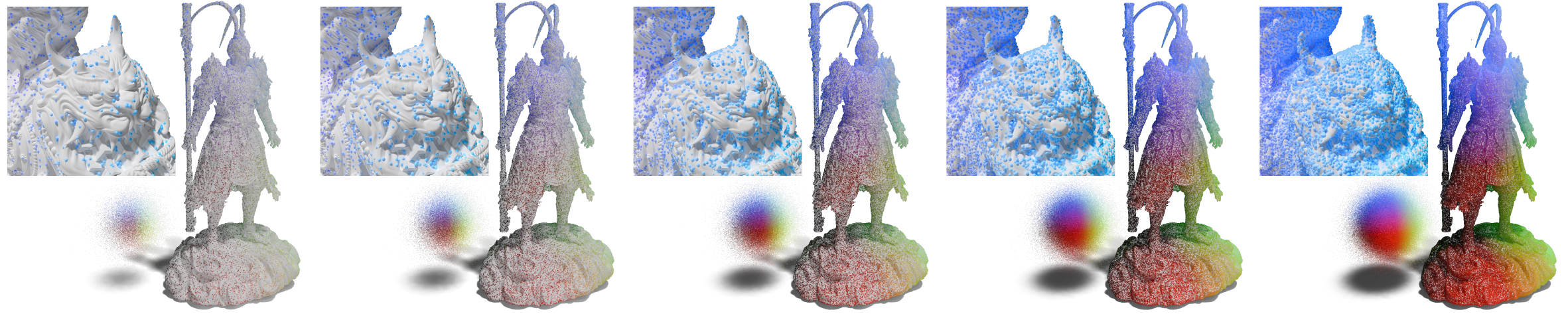}
    \put(1,3){\scriptsize{$n=2^{15}$}}
    \put(21,3){\scriptsize $n=2^{16}$}
    \put(41,3){\scriptsize $n=2^{17}$}
    \put(61,3){\scriptsize $n=2^{18}$}
    \put(81,3){\scriptsize $n=2^{19}$}
    \end{overpic}\vspace{-6pt}
    \caption{\textbf{Forward sampling at different resolutions} on Wukong mesh. For each example, we show the initial Gaussian samples (\emph{bottom left}), the generated samples overlaid on the ground-truth mesh (\emph{right}), and a zoomed-in view (\emph{top left}).}
    \label{fig:res:wukong}
\end{figure*}

\begin{figure*}[t]
    \centering
    \input{tikz/network}\vspace{-20pt}
    \caption{\textbf{Network overview}. \emph{Left}: the training process. \emph{Right}: detailed illustration of the modules. The magnitude-preserving (MP) layers are adpated from~\cite{Karras2024edm2}.}\label{fig:pipeline}\vspace{-3pt}
\end{figure*}

Numerous generative tasks have demonstrated the effectiveness of using diffusion models to learn the mapping from a Gaussian distribution to data distributions.
While previous work is concerned with novel shape synthesis, we are interested in shape representations. Different from existing work, we propose to adapt diffusion models to learn a mapping from a Gaussian distribution to the target distribution of surface points $\Phi_{\gM}$.

Existing networks designed for diffusion models are primarily tailored for regular grids, which have a spatial structure and are high-dimensional. 
There is no straightforward way to adapt existing designs to our setting, \ie, spatial points without regular spatial structure.
A naive idea is to adapt coordinate-based networks (\eg, \cite{park2019deepsdf, muller2022instant}), but they fail to capture detailed geometric features.
For example, \cref{fig:res:ours-vs-mlp} shows the limitations of using standard MLPs and hashing grids to process sampled surface points.
Our network design is inspired by~\cite{Karras2024edm2}, where the inputs and outputs of all layers are standardized to have zero mean and unit variance, resulting in improved performance.
Another key design choice is to resample the training data for each epoch to simulate an infinite number of surface points, approximating the underlying geometry (see \cref{sec:mtd:training}).

\subsection{Inference process: forward \& inverse sampling}\label{sec:mtd:inference}
The mapping between the Gaussian distribution and the surface points distribution is learned via a diffusion model $D_\theta(\cdot, \cdot)$ parameterized by $\theta$. In the literature of diffusion models, $D_\theta(\cdot, \cdot)$ is often called a denoiser. We first discuss the inference process: $\theta$ is known after training, satisfying the ordinary differential equation (ODE):
\begin{equation}\label{eq:ode}
    \mathrm{d}\rvx = \frac{\rvx-D_\theta(\rvx, t)}{t}\mathrm{d}t,
\end{equation}
where $\rvx$ is the 3D position of some sample.

Solving $\rvx(t)$ from \cref{eq:ode} over $t\in[0, T]$ gives the \emph{trajectory} of sample $\rvx$. 
This trajectory connects the Gaussian distribution and the Geometry distribution: $\rvx(0)\sim\Phi_{\gM}$ and $\rvx(T)\sim \gN(\mathbf{0}, T\cdot\mathbf{1})$ i.e., a Gaussian distribution with variance $T$, satisfying 
\begin{equation}\label{eq:normalization}
    \lim_{T\rightarrow\infty} \frac{\rvx(T)}{\sqrt{1 + T^2}}\sim \gN(\mathbf{0}, \mathbf{1}).
\end{equation}
We refer to the sampling process from the Gaussian distribution $\rvx(T)$ to Geometry distribution $\rvx(0)$ as the \emph{forward sampling} (denoted as $\gE$), and the reverse process, from Geometry distribution $\rvx(0)$ to Gaussian distribution $\rvx(T)$, as the \emph{inverse sampling} (denoted as $\gD$). 
The forward and inverse sampling follow the same trajectory but in opposite directions.
In practice we choose discrete timesteps (noise levels) to sample on the trajectory~\cite{karras2022elucidating}, i.e., $T =  t_0>\cdots > t_{i} > t_{i+1} >\dots>t_N =  0$, and denote $\rvx_{i} :=\rvx(t_i)$.

\paragraph{Forward sampling $\gE$.} Starting from $\rvx_0$, a random Gaussian noise, i.e., $\rvx_0 = \rvx(t_0) = T\rvn$ where $\rvn\sim\gN(\mathbf{0}, \mathbf{1})$, we iteratively compute the following steps for $i=0, 1, \cdots, N-1$:
\begin{equation}\label{eq:sampling}
    \rvx_{i+1} = \rvx_{i} + (t_{i+1} - t_{i}) \cdot \frac{\rvx_i - D_\theta(\rvx_i, t_i)}{t_i}, 
\end{equation}
which is an Euler solver for the \cref{eq:ode}.
The endpoint of the trajectory $\rvx_N $ lie on the target surface $\gM$, i.e., $\rvx_N\sim \Phi_{\gM}$.
\cref{eq:sampling} has built a mapping from the \emph{standard Gaussian} distribution $\gN(\mathbf{0}, \mathbf{1})$ to \emph{Geometry} distribution $\Phi_{\gM}$:
if we sample an infinite number of samples from the standard Gaussian distribution, the set of endpoints of their trajectories following \cref{eq:sampling} would closely approximate the surface $\gM$. See~\cref{fig:res:lamp-jellyfish} and~\cref{fig:res:wukong} for some examples.
In practice, we employ a higher-order ODE solver to accelerate the sampling process~\cite{karras2022elucidating}, but for simplicity and clarity, we only show the equations for the simplest case.


 
\begin{algorithm}[b]
\caption{Inverse Sampling}\label{alg:inv-sampling}
\begin{algorithmic}[1]
\Procedure{Inverse Sampling}{$\mathbf{x}$, $t_{i\in\{N, \dots, 0\}}$}
    \State $\rvx_{N}=\rvx$
    \For{$i\in\{N, N-1, \dots, 1\}$}
        \State $\rvd_i = \left(\rvx_i-D_\theta(\rvx_i, t_i)\right)/t_i$
        \State $\rvx_{i-1}=\rvx_{i} + (t_{i-1} - t_i) \cdot \rvd_i$
    \EndFor
    \State $\rvn=\rvx_0/\sqrt{1+t_0^2}$
\EndProcedure
\end{algorithmic}
\end{algorithm}

\paragraph{Inverse sampling $\gD$.}
Starting from a random surface point $\rvx_N \in \gM$, we reverse the trajectory, i.e., iteratively compute for $i = N, N-1, \cdots, 1$: 
\begin{equation}\label{eq:reverse-sampling}
    \rvx_{i-1} = \rvx_{i} + (t_{i-1} - t_{i}) \cdot \frac{\rvx_i - D_\theta(\rvx_i, t_i)}{t_i}.
\end{equation}
The endpoint $\rvx_0$, after normalization $\rvx_0 \leftarrow \nicefrac{\rvx_0}{\sqrt{1+T^2}}$ lies in the noise space according to \cref{eq:normalization}. See~\cref{alg:inv-sampling} for the full algorithm and \cref{fig:res:mouse-inversion} for one example of inverse sampling.
In practice, the inversion process starts from $t_{N}=0$, which causes the denominator in~\cref{eq:reverse-sampling} to be zero. To avoid numerical issues, we instead set $t_N=10^{-8}$.




\subsection{Training process \& network design}\label{sec:mtd:training}
\begin{figure}[t]
    \centering
    \begin{overpic}[trim=0cm 0cm 0cm 0cm,clip,width=1\linewidth,grid=false]{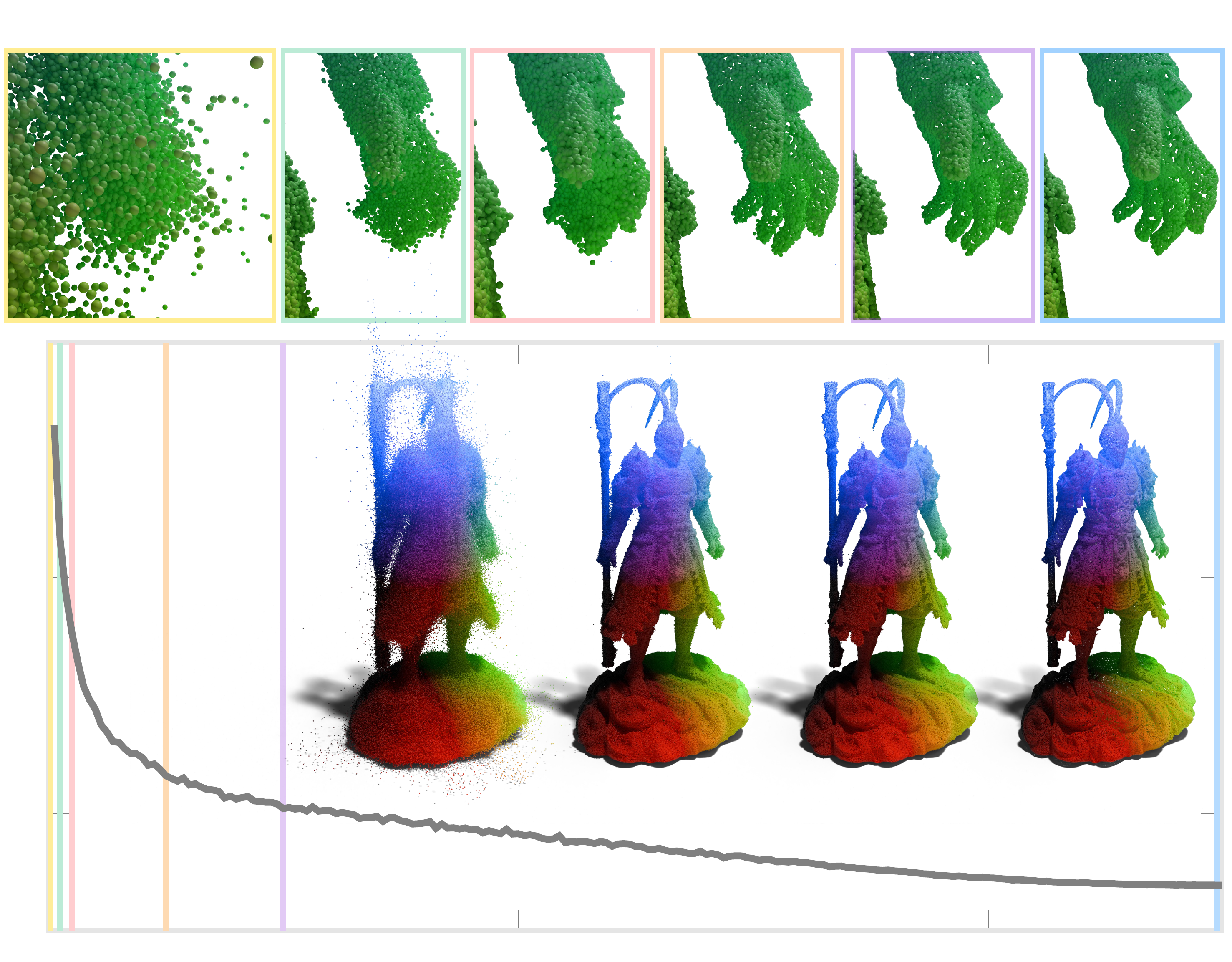}
    \put(8,75){\scriptsize $k=0$}
    \put(25,75){\scriptsize $k=10$}
    \put(41,75){\scriptsize $k=20$}
    \put(55,75){\scriptsize $k=100$}
    \put(71,75){\scriptsize $k=200$}
    \put(86,75){\scriptsize $k=1000$}
    \put(-3,11){\tiny $0.003$}
    \put(-3,30){\tiny $0.004$}
    \put(-3,49){\tiny $0.005$}
    \put(3,0){\tiny 0}
    \put(96,0){\tiny 1000}
    \put(32,13){\scriptsize $k=0$}
    \put(50,13){\scriptsize $k=10$}
    \put(68,13){\scriptsize $k=20$}
    \put(83,13){\scriptsize $k=1000$}
    \put(45,-0.5){\scriptsize\itshape epochs}
    \end{overpic}\vspace{-3pt}
    \caption{\textbf{Training process}. We show the Chamfer distance over epochs and highlight intermediate results (\emph{bottom}). By the $10$-th epoch, the network already captures the overall geometry, with finer details further refined in later iterations, as seen in the zoomed-in hand region (\emph{top}).}
    \label{fig:res:training-wukong}
\end{figure}

Given the input geometry $\gM$, we first generate the training set by sampling a set of surface points $\{\rvx\in\gM\}$. 
Following~\cite{karras2022elucidating}, we add noise to the data $\rvy=\rvx+\sigma\rvn$ where $\sigma$ indicates the noise level, and optimize the denoiser network:
\begin{equation}\label{eq:training}
\argmin_\theta\E_{\rvx\in\gM}\E_{\rvn\sim\gN(\mathbf{0}, \mathbf{1})}\E_{\sigma>0}\|D_\theta(\rvx+\sigma\rvn, \sigma) - \rvx\|,
\end{equation}
Our network is simple yet effective: the noise levels $\sigma$, standard Gaussian noise $\rvn$, and input coordinates $\{\rvx\}$ are projected to high-dimensional space following~\cite{zhang20223dilg}. See~\cref{fig:pipeline} for full details of our network design and \cref{fig:res:training-wukong} for an example of the training process.

Recall that our goal is to have the learned geometry distribution to accurately approximate the target surface from an infinite number of Gaussian samples. 
To simulate this, we require a training dataset with an infinite number of surface points.
In practice, we \emph{resample} a set of $2^{25}$ surface points for training before each epoch. 
Over 1000 epochs, the network encounters a sufficiently large number of ground-truth surface points. 
This approach is fundamentally different from typical deep learning applications, where the training set is preprocessed and fixed prior to training. 
In our setting, however, the training datasets—\ie, surface points—are intentionally varied across epochs.

\section{Experiments}

\subsection{Implementation}
The code is implemented with PyTorch.
For most experiments, we use 6 blocks and $C=512$ for all linear layers, resulting in 5.53 million parameters.
One epoch (512 iterations) of training takes approximately 2.5 minutes to complete on 4 A100 GPUs. 
Training typically requires several hours to achieve reasonably good results.
\cref{fig:res:training-wukong} shows one example of training quality over epochs.


To quantify the accuracy of our approach, we measure the distance between samples from our Geometry distribution, $\gX_{\text{gen}}$, and the ground-truth surface $\gM$. 
Specifically, we sample 1 million surface points from $\gM$, denoted as $\gX_{\text{ref}}$, as the reference set. We then compute the Chamfer distance between the two sets, $\gX_{\text{gen}}$ and $\gX_{\text{ref}}$, as our metric.

In the following we will investigate multiple applications of our novel shape representations, ablate our design choices and verify the correctness of the inversion.

\subsection{Applications}\label{sec:app}

\begin{figure}[t]
    \centering
    \begin{overpic}[trim=0cm 0cm 0cm 0cm,clip,width=1\linewidth,grid=false]{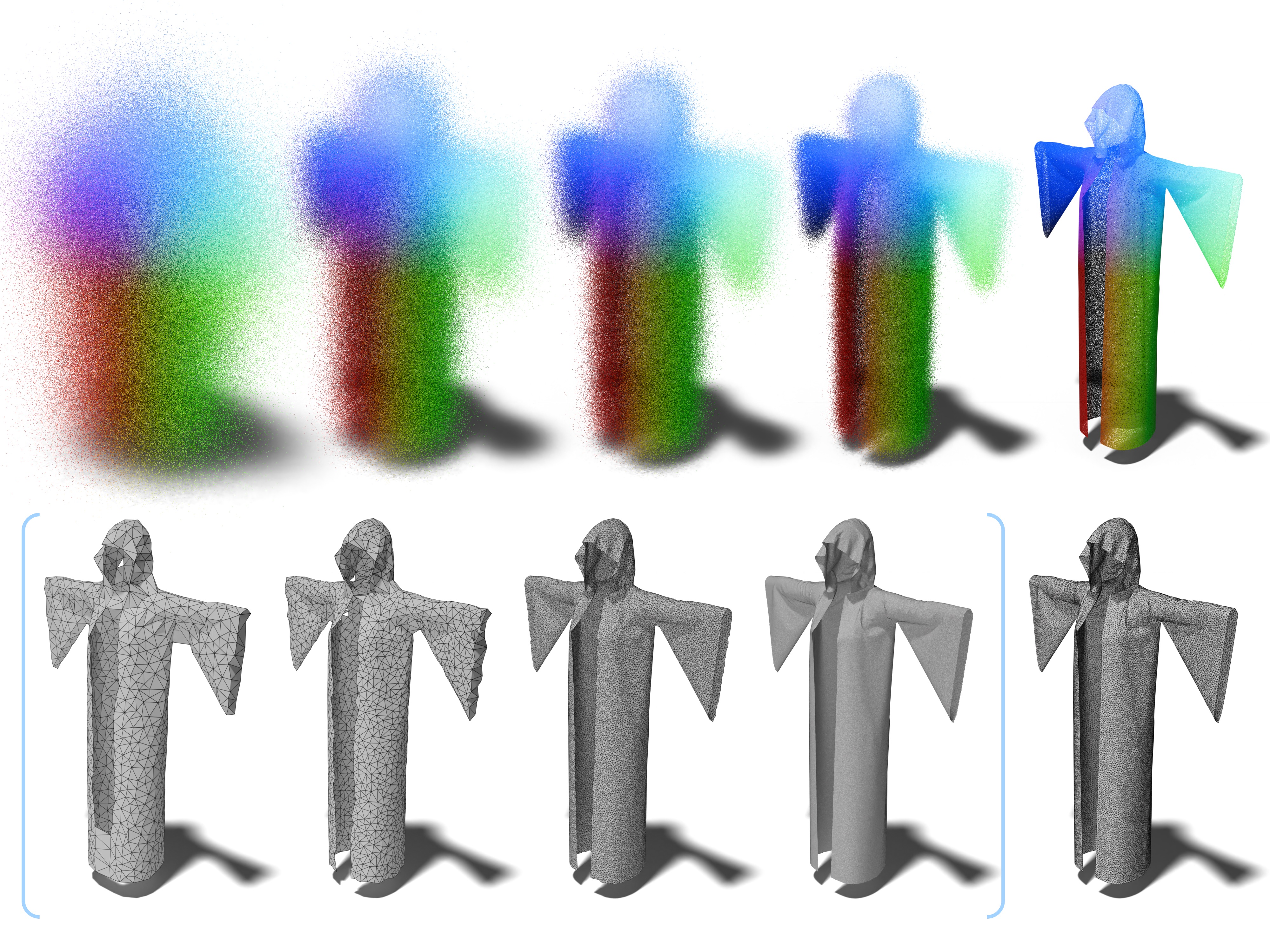}
    \put(6,35){\scriptsize $n=1$K}
    \put(24,35){\scriptsize $n=2$K}
    \put(43,35){\scriptsize $n=20$K}
    \put(61,35){\scriptsize $n=200$K}
    \put(12,2){\scriptsize reconstructed mesh in different resolution $n$}
    \put(82,2){\scriptsize ground-truth}
    \put(8,72){\scriptsize $i=40$}
    \put(28,72){\scriptsize $i=45$}
    \put(47,72){\scriptsize $i=47$}
    \put(65,72){\scriptsize $i=50$}
    \put(84,72){\scriptsize $i=64$}
    \end{overpic}\vspace{-6pt}
    \caption{\textbf{Application: remeshing}. \emph{Top}: starting from a Gaussian distribution, we show the intermediate steps at different timesteps $t$. 
    \emph{Bottom}: we use Ball Pivoting to reconstruct a mesh. The number $n$ indicates the number of points used in the reconstruction. Since our method supports infinitely many points sampling, we show results obtained using different number $n$. The more points we have, the better we can approximate the original surface. The mesh is taken from~\cite{korosteleva2021generating}.}
    \label{fig:res:jacket}
\end{figure}

We can generate a varying number of samples from the geometry distribution for surface remeshing at different resolutions.
In~\cref{fig:res:jacket}, we use the Ball Pivoting algorithm~\cite{bernardini1999ball}, implemented in MeshLab~\cite{meshlab}, with default parameters, to triangulate the samples at different resolutions. 
Note that this example also illustrates the effectiveness of our method in representing non-watertight surfaces, where most implicit function-based methods would fail.

\begin{figure*}[t]
\begin{overpic}[trim=0cm 0cm 0cm 0cm,clip,width=1\linewidth,grid=false]{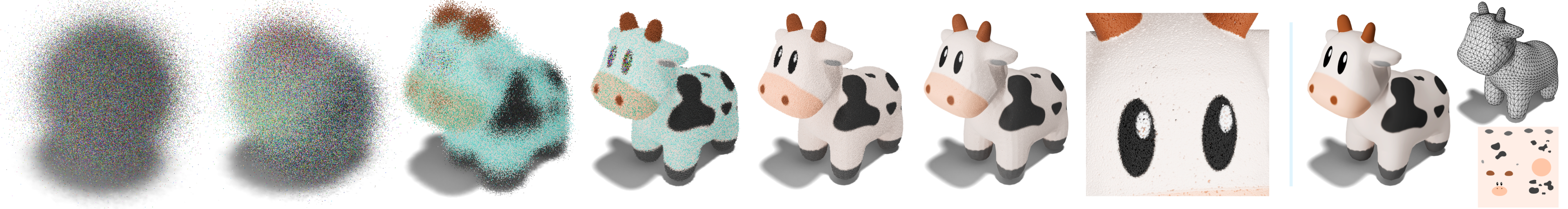}
\put(5,13.5){\scriptsize $i=0$}
\put(17,13.5){\scriptsize $i=40$}
\put(29,13.5){\scriptsize $i=48$}
\put(40,13.5){\scriptsize $i=56$}
\put(51,13.5){\scriptsize $i=60$}
\put(61,13.5){\scriptsize $i=64$}
\put(70,13.5){\scriptsize\itshape zoom-in at $i=64$}
\put(86,13.5){\scriptsize\itshape ground-truth}
\end{overpic}\vspace{-6pt}
\caption{\textbf{Application: textured geometry.} The proposed representation can also be used for textured geometry. \emph{Left}: 1 million points with texture (6-dimensional vectors) at different timesteps $t$. \emph{Right}: the ground-truth geometry and texture.}
\label{fig:res:spot-texture}
\end{figure*}


\begin{figure}[t]
    \centering
    \begin{overpic}[trim=0cm 0cm 0cm 0cm,clip,width=1\linewidth,grid=false]{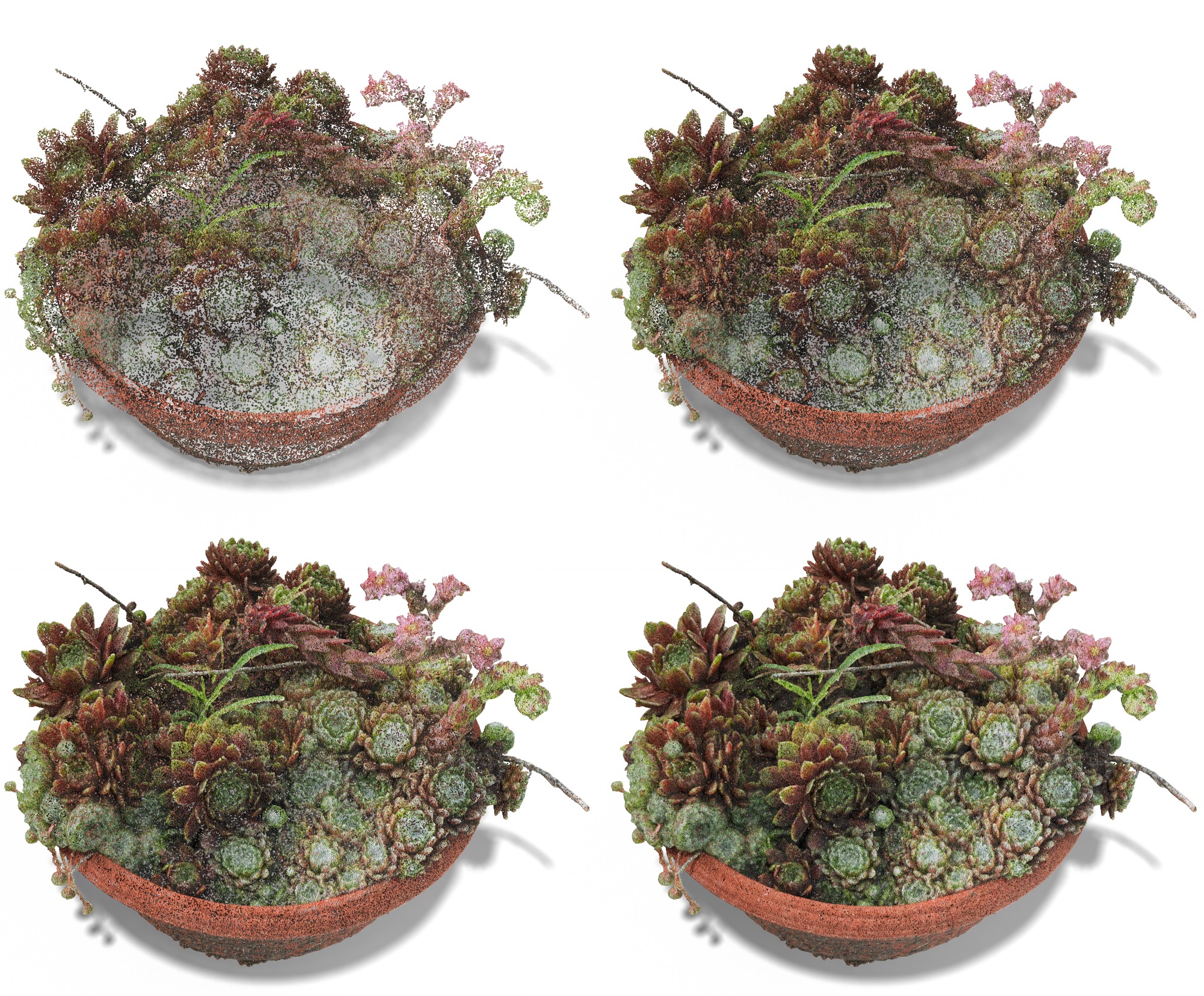}
    \put(15,80){\scriptsize $n=250K$}
    \put(65,80){\scriptsize $n=500K$}
    \put(13,39){\scriptsize $n=1,000K$}
    \put(63,39){\scriptsize $n=2,000K$}
    \end{overpic}\vspace{-6pt}
    \caption{\textbf{Application: combination with color field network.} We show results of different numbers of points.}
    \label{fig:res:houseleek}
\end{figure}

\begin{figure}[t]
    \centering
    \includegraphics[width=1\linewidth]{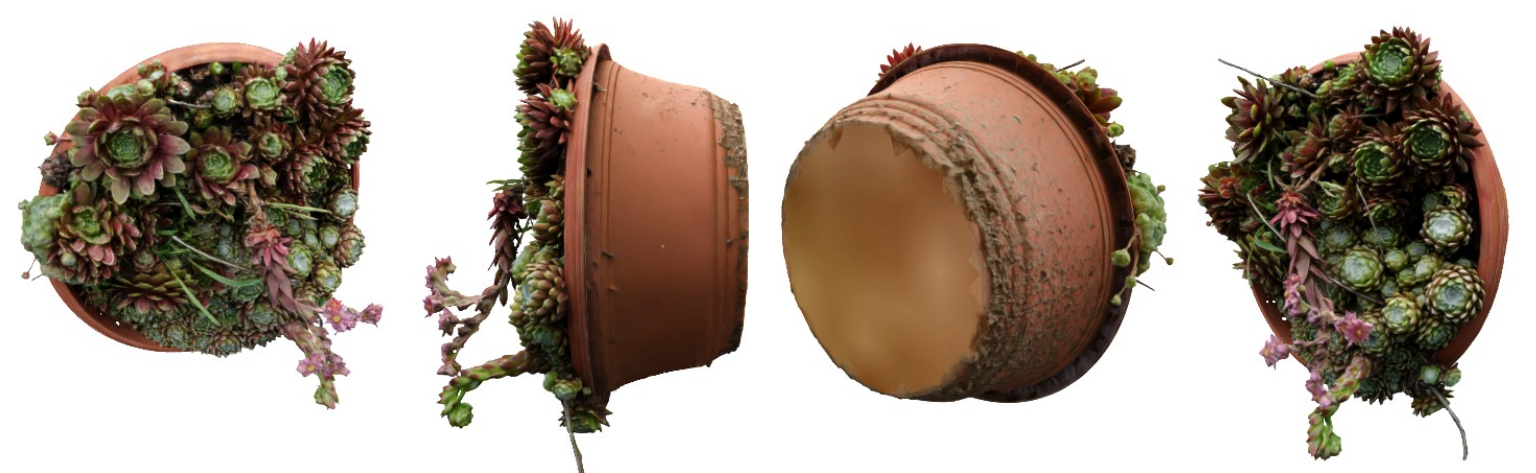}
    \vspace{-20pt}
    \caption{\textbf{Application: photo-realistic rendering with Gaussian splatting.} These views are not visible during training.}
    \label{fig:res:gs}
\end{figure}

Geometry distributions can be further extended to incorporate additional information such as color or motion. 
\cref{fig:res:spot-texture} shows an example of feeding the texture color in addition to the 3D position of each surface point during training (i.e., the $\rvx$ in~\cref{eq:training} is 6-dim).
\cref{fig:res:houseleek} shows an alternative approach: a separate color field network based on hashing grids~\cite{muller2022instant} is trained, allowing the querying of color vectors for all spatial points. 
See more results of textured geometry distributions in the supplementary materials.

Furthermore, the sampled points can serve as inputs for Gaussian splatting~\cite{kerbl20233d}, as shown in~\cref{fig:res:gs}. 
Specifically, we sample 1 million points from the distribution to initialize the Gaussian splatting, disabling point gradients and point pruning in the original implementation during training. This optimization assigns colors, radius, and scaling to the points, and can be used for novel view synthesis.

\begin{figure}[t]
    \centering
    \includegraphics[width=1\linewidth]{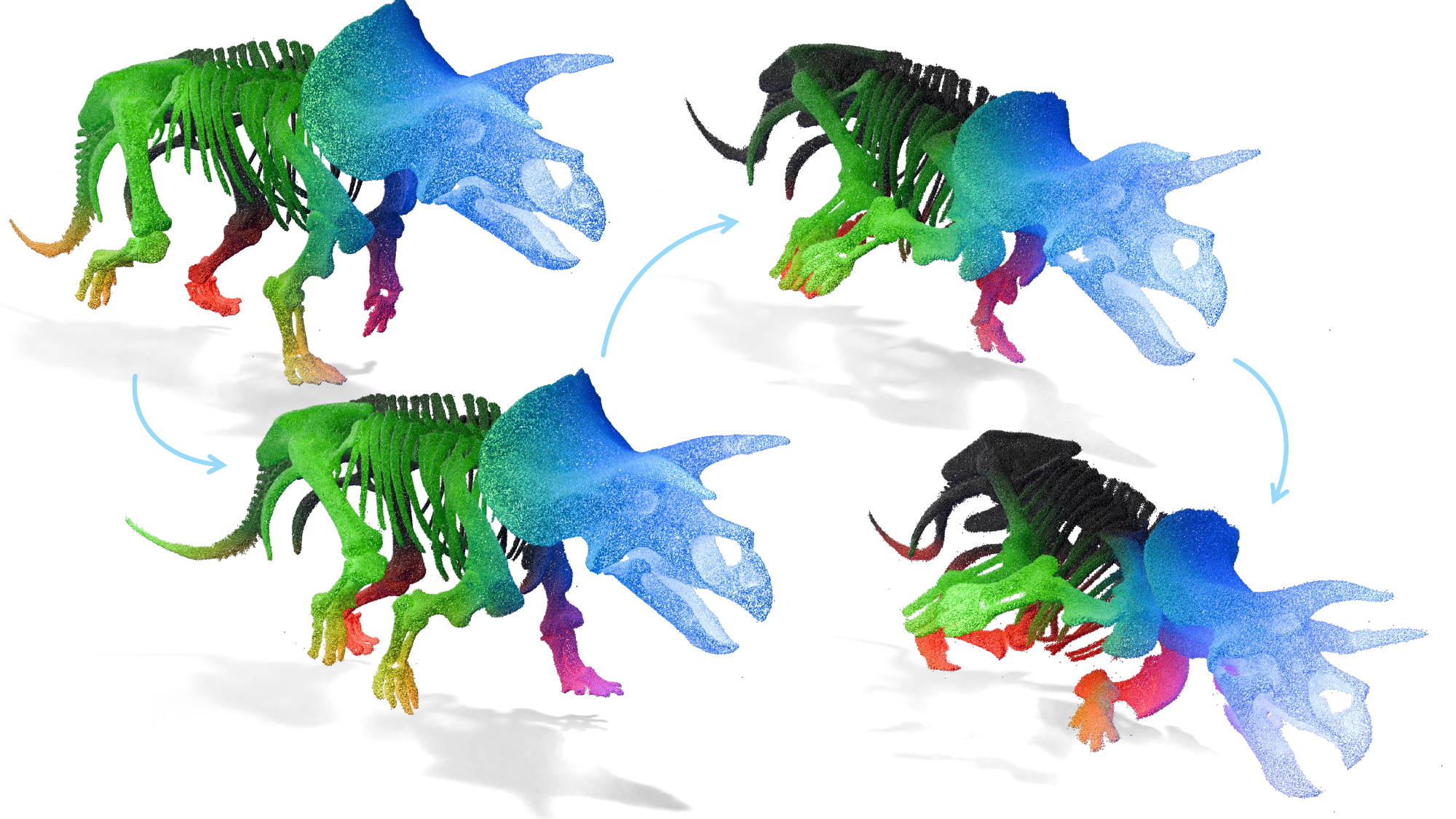}
    \vspace{-20pt}
    \caption{\textbf{Application: dynamic object modeling.} We use a single network to learn the motion of the geometry distribution. Only 4 out of 250 frames are shown here.}
    \label{fig:res:dynamic}
\end{figure}

Finally, we show an extension to dynamic geometries (4D objects), achieved by adding a temporal input to the denoiser network $D_\theta$, making the inputs 4D. 
The trained network encodes the motions of the geometry distributions. See one example in~\cref{fig:res:dynamic}.





\begin{table}[t]
    \centering\hfill
    \begin{subtable}{0.5\linewidth}\centering
        \input{tables/tab_network_arch}
        \caption{tested on Loong shape}\label{tab:ablation:arch}
    \end{subtable}\hfill%
    \begin{subtable}{0.5\linewidth}
        \input{tables/tab_ablation_data_size}
        \caption{tested on jellyfish shape}\label{tab:ablation:data-size} 
    \end{subtable}\hfill
    \begin{subtable}{0.5\linewidth}\centering
        \input{tables/tab_sampling_steps}
        \caption{tested on Archimedes shape}\label{tab:ablation:sampling-steps} 
    \end{subtable}\hfill
    \begin{subtable}{0.5\linewidth}
        \input{tables/tab_ablation_block_size}
        \caption{tested on lamp shape}\label{tab:ablation:block-size} 
    \end{subtable}\hfill\vspace{-6pt}
\caption{\textbf{Ablation studies} on network architecture, dataset size, sampling steps, and network blocks, tested on shapes from~\cref{fig:teaser}. }\label{tab:ablation}    
\end{table}

\begin{figure*}[t]
    \centering
    \begin{overpic}[trim=0cm 0cm 2cm -0.5cm,clip,width=1\linewidth,grid=false]{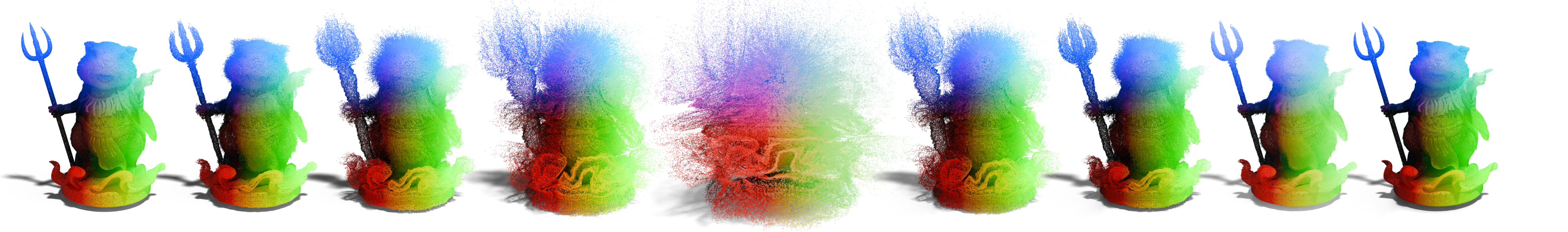}
    \put(0,15){\vector(1,0){49}}
    \put(51,15){\vector(1,0){49}}
    \put(20, 16){\scriptsize Inverse Sampling $\gD$}
    \put(71, 16){\scriptsize Sampling $\gE$}
    \end{overpic}
    \caption{Inverse sampling $\gD$ and sampling $\gE$ for 1M points. Both inverse sampling and sampling are using $N=64$ steps. Note that, the image in the middle is the noise space, where it does not look like a Gaussian distribution. This implies that the mapping is not bijective. Some points in the noise space will never be mapped to from the shape space.
    }
    \label{fig:res:mouse-inversion}
\end{figure*}

\subsection{Ablation studies}\label{sec:exp:mlp}
Using distributions to model a surface shows advantages over vector field-based methods~\cite{chibane2020neural, yang2023neural, zhou2024cap} which usually fail to produce uniform sampling: as shown in~\cref{fig:res:lion-vf-ours}, even at extremely high resolutions with 1 million samples, their samples fail to adequately cover the target surface. More results can be found in the supplementary materials.

As mentioned earlier, adapting well-established diffusion models from tasks involving regular grid data to our setting, which focuses on learning geometry distributions, may seem straightforward but proves challenging. 
We compare our network with two established architectures. 
The first is hashing grids~\cite{muller2022instant}, originally designed for volume rendering with 3-dimensional coordinate inputs. 
We adapt it to accept 4-dimensional inputs (3 for coordinates and 1 for noise level). 
The second is a simple MLP network based on DeepSDF~\cite{park2019deepsdf}, where we concatenate point and noise level embeddings as inputs.
As shown in~\cref{tab:ablation:arch}, our proposed network significantly outperforms these straightforward adaptions.
While the Chamfer distance for the MLP-baseline appears promising, the qualitative results in \cref{fig:res:ours-vs-mlp} reveal that this baseline fails to capture fine details.

Consistent with observations in other diffusion models, we find that a larger training set, more sampling steps, and deeper networks lead to higher accuracy and improved generation quality. We provide ablation studies in~\cref{tab:ablation} to validate these findings.
Moreover, although both types of distributions work effectively, we observe that the Gaussian distribution performs slightly better than the uniform distribution in most cases, as shown in~\cref{tab:gaussian-uniform}.

\begin{table}[t]
\input{tables/tab_gaussian_uniform}\vspace{-4pt}
\caption{\textbf{Ablation study} on using Uniform and Gaussian distributions as initial noise sources. We report
Chamfer distance ($\times 10^3$) on different shapes from~\cref{fig:teaser}.
}\label{tab:gaussian-uniform} 
\end{table}

\begin{table}[t]
\input{tables/tab_inversion_steps}\vspace{-4pt}
\caption{\textbf{Mean squared error} of $\Vert \vec{x} - \gE\circ \gD(\vec{x}) \Vert_2^2$ with different inversion steps, evaluated on the mouse shape in \cref{fig:res:mouse-inversion}.
}\label{tab:inverison-steps} 
\end{table}

\subsection{Inversion}\label{sec:res:inversion}
\begin{figure}[t]
    \centering
    \begin{overpic}[trim=0cm 0cm 0cm -1.4cm,clip,width=1\linewidth,grid=false]{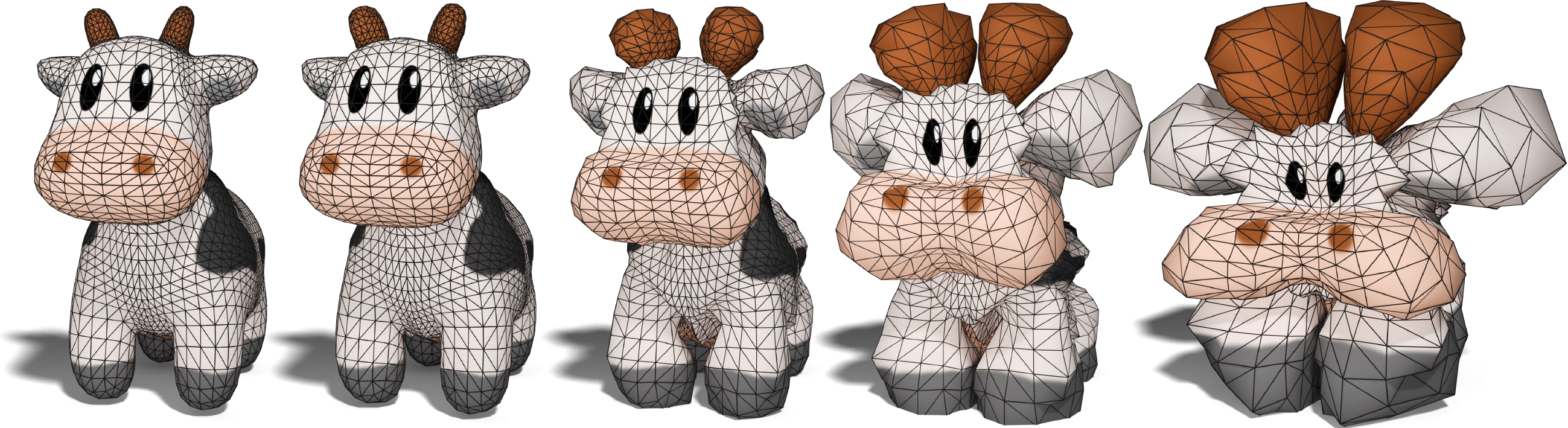}
    \put(4.6,28){\scriptsize $i=0$}

    \put(22,28){\scriptsize $i=8$}

    \put(40.6,28){\scriptsize $i=15$}

    \put(58.6,28){\scriptsize $i=20$}

    \put(81.6,28){\scriptsize $i=64$}
    \end{overpic}
    \caption{We use the inversion $\gD(\cdot)$ to map the spot mesh back to the noise space. Only the original mesh vertices are mapped; textures shown here are for correspondence purposes only.}
    \label{fig:res:spot-inversion}
\end{figure}

As discussed in~\cref{sec:mtd:inference}, our network learns the trajectory connecting the Gaussian distribution and the Geometry distribution. 
The forward and inverse sampling follow this trajectory in opposite directions. 
In other words, for a surface point-\ie, a sample drawn from Geometry distribution $\rvx\sim\Phi_{\gM}$-the composition of inverse and forward sampling applied to this sample should also follow the Geometry distribution: $\gE\circ\gD(\rvx)\sim\Phi_{\gM}$.
To validate this, we sample 1 million surface points, denoted as $\{\rvx\}$, and apply inverse sampling, following~\cref{eq:reverse-sampling}, to obtain a set of Gaussian noise samples $\{\gD(\rvx)\}$. 
We then apply forward sampling on $\{\gD(\rvx)\}$, following~\cref{eq:sampling}, to obtain $\{\gE\circ\gD(\rvx)\}$. 
Finally, we evaluate the mean squared error (MSE) between $\{\rvx\}$ and $\{\gE\circ\gD(\rvx)\}$, as they are in one-to-one correspondence.
In~\cref{fig:res:mouse-inversion} we show intermediate results from the inverse and forward sampling. We can see that indeed both the initial samples (leftmost) and the results after composition (rightmost) align with the Geometry distribution. 
\cref{tab:inverison-steps} reports the MSE for different choices of inversion steps.
In~\cref{fig:res:spot-inversion}, we apply inverse sampling to the original surface vertices with the ground-truth triangulation and texture coordinates, to demonstrate that our inversion is semantically meaningful.
In the supplementary materials we show additional interesting results: we composite inverse sampling and forward sampling from \emph{different} surfaces, yet still obtain expected results. This further demonstrates the validity of our method.

\section{Conclusion}
We have introduced a novel geometric data representation that addresses key limitations of traditional methods, such as watertightness and manifold constraints. 
Our approach models 3D surfaces as geometry distributions encoded in a diffusion model, allowing flexible and precise sampling on complex geometries.
This work advances neural 3D representation techniques and establishes a foundation for further exploration and development in geometry modeling, processing, and analysis.

As a first attempt in this field, there are many exciting avenues for future research. We just highlight selected examples but hope that our initial presentation motivates others to explore this shape representation. First, the training of diffusion models builds a trajectory between the Gaussian distribution and the geometry distribution, which can be interpreted as a mapping between two distributions. We propose to investigate how to incorporate regularizers into this mapping during training, such as area/volume preservation or semantic meaningfulness. 
Second, we are also interested in exploring how to define neural geometry operators on geometry distributions, similar to the well-investigated geometry processing operators on triangle meshes~\cite{botsch2010polygon,aigerman2022neural}. Third, we have shown some preliminary meshing results in~\cref{fig:res:jacket}. However, meshing is generally a challenging problem requiring precise algorithms to convert spatial data into graphs (vertices and faces). An interesting avenue of research is to investigate joint sampling and meshing algorithms for the proposed representation.



\clearpage

\section*{Acknowledgements}
This work was supported by funding from King Abdullah University of Science and Technology (KAUST) — Center of Excellence for Generative AI, under award number 5940.


{
    \small
    \bibliographystyle{ieeenat_fullname}
    \bibliography{bib}
}




\end{document}

%% file: preamble.tex
\input{math_commands.tex}

\usepackage{booktabs}
\usepackage{xspace}
\usepackage{comment}
\usepackage{overpic}

\usepackage{algorithm}
\usepackage{algpseudocode}

\usepackage{comment}

\usepackage{tikz}
\usepackage{pgfplots}
\usetikzlibrary{positioning,calc,fadings,backgrounds,fit,3d,shapes.misc,shapes.geometric,matrix,hobby}
\pgfplotsset{compat=1.14}
\usepackage{wrapfig}
\usepackage[fixed]{fontawesome5}
\usepackage{colortbl}
\usepackage{xcolor}

\usepackage{wrapfig}

\definecolor{tabcol}{HTML}{a2d2ff}
\definecolor{mygreen}{HTML}{57cc99}
\definecolor{myred}{HTML}{ff595e}
\definecolor{mypurple}{HTML}{9d4edd}
\definecolor{myyellow}{HTML}{ffd100}
\definecolor{myorange}{HTML}{fb8500}
\definecolor{mydarkblue}{HTML}{5aa9e6}
\definecolor{mybrown}{HTML}{754632}
\definecolor{textblue}{HTML}{42a5f5} 
\newcommand{\mytabtext}[1]{{\itshape\bfseries #1}}

\newcommand\coolname{\textcolor{black}{\sc\textbf{Geometry Distributions}}\xspace}
\newcommand\coolnameshort{\textcolor{black}{\sc\textbf{GeomDist}}\xspace}

\usepackage{pifont}
\newcommand{\cmark}{\textcolor{black}{\ding{51}}}%
\newcommand{\xmark}{{\textcolor{myred}{\ding{55}}}}%

\renewcommand{\vec}[1]{\mathbf{#1}} 

\usepackage{multirow}
\usepackage{nicefrac}

\usepackage{listings}
\definecolor{codegreen}{rgb}{0,0.6,0}
\definecolor{codegray}{rgb}{0.5,0.5,0.5}
\definecolor{codepurple}{rgb}{0.58,0,0.82}
\definecolor{backcolour}{rgb}{0.95,0.95,0.92}

\lstdefinestyle{mystyle}{
    backgroundcolor=\color{backcolour},   
    commentstyle=\color{codegreen},
    keywordstyle=\color{magenta},
    numberstyle=\tiny\color{codegray},
    stringstyle=\color{codepurple},
    basicstyle=\ttfamily\footnotesize,
    breakatwhitespace=false,         
    breaklines=true,                 
    captionpos=b,                    
    keepspaces=true,                 
    numbers=left,                    
    numbersep=5pt,                  
    showspaces=false,                
    showstringspaces=false,
    showtabs=false,                  
    tabsize=2
}

%% file: math_commands.tex
\usepackage{amsmath,amsfonts,bm}

\def\eqref#1{equation~\ref{#1}}

\def\1{\bm{1}}

\def\rvd{{\mathbf{d}}}

\def\rvf{{\mathbf{f}}}

\def\rvn{{\mathbf{n}}}

\def\rvp{{\mathbf{p}}}

\def\rvv{{\mathbf{v}}}

\def\rvx{{\mathbf{x}}}
\def\rvy{{\mathbf{y}}}

\DeclareMathAlphabet{\mathsfit}{\encodingdefault}{\sfdefault}{m}{sl}
\SetMathAlphabet{\mathsfit}{bold}{\encodingdefault}{\sfdefault}{bx}{n}

\def\gD{{\mathcal{D}}}
\def\gE{{\mathcal{E}}}
\def\gF{{\mathcal{F}}}

\def\gM{{\mathcal{M}}}
\def\gN{{\mathcal{N}}}

\def\gP{{\mathcal{P}}}

\def\gV{{\mathcal{V}}}

\def\gX{{\mathcal{X}}}

\def\sR{{\mathbb{R}}}

\newcommand{\E}{\mathbb{E}}

\DeclareMathOperator*{\argmin}{arg\,min}

%% file: tables/tab_data_representation.tex
\centering \footnotesize
    {\def\arraystretch{1.2}\tabcolsep=0.4em 
    \begin{tabular}{cccccc}
    \toprule[1pt]
     \mytabtext{types} &  \mytabtext{structure} & \mytabtext{infinity} & \mytabtext{surface} & \begin{tabular}[c]{@{}c@{}}\mytabtext{non-}\\ \mytabtext{watertight}\end{tabular} & \begin{tabular}[c]{@{}c@{}} \mytabtext{color/} \\ \mytabtext{textures}\end{tabular}\\
    \midrule[0.5pt]
    
    \mytabtext{point clouds} &  $\{\rvp_i\in\sR^3\}_{i\in\gP}$ & \xmark & \xmark & \cmark & \cmark\\

    \rowcolor{tabcol!30} 
    \mytabtext{meshes} &  \begin{tabular}[c]{@{}c@{}} $\big(\{\rvv_i\in\sR^3\}_{i\in\gV}$,\\  $ \{\rvf_j\in\gV^3\}_{j\in\gF} \big)$\end{tabular} & \xmark & \cmark & \cmark & \cmark\\

    \mytabtext{voxels} & $\sR^{d\times h\times w\times c}$ & \xmark & \cmark & \cmark & \cmark\\
    
    \rowcolor{tabcol!30} 
    \mytabtext{SDFs} &  $\{\rvp|f(\rvp)=0\}$  & \cmark & \cmark & \xmark & \xmark\\
    \midrule[0.5pt]
    \coolnameshort &  $\{\gE(\rvn)\}$ & \cmark & \cmark  & \cmark & \cmark\\
    \bottomrule[1pt]
    \end{tabular}
    }

%% file: tikz/network.tex
\resizebox{\linewidth}{!}{%

    \begin{tikzpicture}
        \definecolor{dandelion}{rgb}{0.94, 0.88, 0.19}
        \definecolor{columbiablue}{rgb}{0.61, 0.87, 1.0}
                
        \definecolor{darkcoral}{HTML}{73234a}

        \tikzstyle{arrow} = [thick,->,>=stealth]

        \tikzstyle{network} = [
            rectangle, fill=lightgray!15,
        ]

        \tikzstyle{mpfourier} = [
            rectangle, draw=darkcoral!80, fill=darkcoral!10, minimum width=2cm,
        ]

        \tikzstyle{mplinear} = [
            rectangle, draw=mygreen!80, fill=mygreen!10, minimum width=2cm,
        ]

        \tikzstyle{mpsilu} = [
            rectangle, draw=myred!80, fill=myred!10, minimum width=2cm,
        ]

        \tikzstyle{pointembed} = [
            rectangle, draw=myyellow!80, fill=myyellow!10, minimum width=2cm,
        ]

        \tikzstyle{normalize} = [
            rectangle, draw=mybrown!80, fill=mybrown!10, minimum width=2cm,
        ]

        \tikzstyle{mpadd} = [
            rectangle, draw=myorange!80, fill=myorange!10, minimum width=2cm,
        ]

        \tikzstyle{block} = [
            rectangle, draw=tabcol!80, fill=tabcol!20, minimum height=0.3cm,
        ]

        \tikzstyle{final} = [
            rectangle, draw=mypurple!60, fill=mypurple!10, minimum height=0.3cm,
        ]

        \tikzstyle{levelemb} = [
            rectangle, draw=brown!60, fill=brown!10, minimum height=0.3cm,
        ]

        \tikzstyle{inputemb} = [
            rectangle, draw=magenta!60, fill=magenta!10, minimum height=0.3cm,
        ]

        \node[] (noise_level_denoiser) {Level};
        \node[inner sep=0pt, below=0.3cm of noise_level_denoiser] (noise_level_size_denoiser) {$B\times 1$};

        \node[right=0.3cm of noise_level_denoiser] (noise_denoiser) {Noise};
        \node[inner sep=0pt, ] at (noise_level_size_denoiser -| noise_denoiser) (noise_size_denoiser) {$B\times 3$};

        \node[right=0.3cm of noise_denoiser] (mesh_denoiser) {Mesh};
        \node[inner sep=0pt, ] at (noise_size_denoiser -| mesh_denoiser) (mesh_size_denoiser) {$B\times 3$};
        \node[inner sep=0pt, below=0.3cm of mesh_size_denoiser] (plus_denoiser) {$\bigoplus$};

        \node[inner sep=0pt] at (plus_denoiser -| noise_size_denoiser) (times_denoiser) {$\bigotimes$};

        \coordinate[] (below_noise_level_size_denoiser) at (times_denoiser -| noise_level_size_denoiser);
        
        \draw[arrow] (noise_level_denoiser) -- (noise_level_size_denoiser);
        \draw[arrow] (noise_denoiser) -- (noise_size_denoiser);
        \draw[arrow] (mesh_denoiser) -- (mesh_size_denoiser);
        \draw[arrow] (noise_size_denoiser) -- (times_denoiser);
        \draw[arrow] (mesh_size_denoiser) -- (plus_denoiser);
        \draw[arrow] (below_noise_level_size_denoiser) -- (times_denoiser);
        \draw[arrow] (times_denoiser) -- (plus_denoiser);

        \node[inputemb, below=0.3cm of plus_denoiser] (inputemb_denoiser) {\textcolor{magenta}{\bfseries\textsc{Input Emb}}};
        \node[levelemb, ] at (inputemb_denoiser -| noise_level_size_denoiser) (levelemb_denoiser) {\textcolor{brown}{\bfseries\textsc{Level Emb}}};


        \draw[arrow] (plus_denoiser) -- (inputemb_denoiser);
        \draw[arrow] (noise_level_size_denoiser) -- (levelemb_denoiser);

        \begin{pgfonlayer}{background}
            \node [fit=(times_denoiser)(plus_denoiser)(inputemb_denoiser)] (above_network_denoiser) {}; 
        \end{pgfonlayer}

        \node[
            block, 
            fit=(noise_size_denoiser)(mesh_size_denoiser),
            below=0.3cm of above_network_denoiser,
            text height=1.5ex,
        ] (block1) {\textcolor{textblue}{\bfseries\textsc{Mid Block}}};


        \node[inner sep=0pt, below=0.0cm of block1] (vdots_denoiser) {$\vdots$};

        \node[
            block, 
            fit=(noise_size_denoiser)(mesh_size_denoiser),
            below=0.2cm of vdots_denoiser,
            text height=1.5ex,
        ] (block3) {\textcolor{textblue}{\bfseries\textsc{Mid Block}}};

        \node[
            final, 
            fit=(noise_size_denoiser)(mesh_size_denoiser),
            below=0.0cm of block3,
            text height=1.5ex,
        ] (final) {\textcolor{mypurple}{\bfseries\textsc{Final Block}}};

        \begin{pgfonlayer}{background}
            \node [network, fit=(block1)(block3)(final)(levelemb_denoiser)] {}; 
        \end{pgfonlayer}


        
        \draw[arrow] (inputemb_denoiser) -- (block1.north -| inputemb_denoiser);

        \draw[arrow] (levelemb_denoiser) |- (block1.west);
        \draw[arrow] (levelemb_denoiser) |- (block3.west);
        \draw[arrow] (levelemb_denoiser) |- (final.west);

        \node[below=0.5cm of final] (output_denoiser) {$B\times 3$};
        \draw[arrow] (final) -- (output_denoiser);
    
        \begin{pgfonlayer}{background}
            \node [fit=(noise_level_denoiser)(noise_denoiser)(mesh_denoiser), label={above:\bfseries\textsc{Training}}] {}; 
        \end{pgfonlayer}

        \node[right=2.0cm of mesh_denoiser] (noise_emb) {Level};

        \node[mpfourier, below=0.5cm of noise_emb] (mpfourier_emb) {MPFourier};
        \node[mplinear, below=0cm of mpfourier_emb] (mplinear_emb) {MPLinear};
        \node[mpsilu, below=0cm of mplinear_emb] (mpsilu_emb) {MPSiLU};

        \draw[arrow] (noise_emb) -- (mpfourier_emb);

        \node[below=0.5cm of mpsilu_emb] (emb_emb) {LevelEmb};
        \draw[arrow] (mpsilu_emb) -- (emb_emb);

        \node[below=0.5cm of emb_emb] (output_emb) {To all blocks};
        \draw[arrow] (emb_emb) -- (output_emb);

        \begin{pgfonlayer}{background}
            \node [fit=(noise_emb), label={above:\textcolor{brown}{\bfseries\textsc{Level Embed}}}] {}; 
        \end{pgfonlayer}

        \begin{pgfonlayer}{background}
            \node [levelemb, fit=(mpfourier_emb)(mpsilu_emb),] {}; 
        \end{pgfonlayer}

        \node[right=2.0cm of noise_emb] (noise_input) {Input};

        \node[pointembed, below=0.5cm of noise_input] (pointembed_input) {PointEmb};
        \node[mplinear, below=0cm of pointembed_input] (mplinear_input) {MPLinear};
        \node[below=0.5cm of mplinear_input] (emb_input) {InputEmb};

        \draw[arrow] (noise_input) -- (pointembed_input);
        \draw[arrow] (mplinear_input) -- (emb_input);

        \node[below=0.5cm of emb_input] (output_input) {To block$_0$};
        \draw[arrow] (emb_input) -- (output_input);

        \begin{pgfonlayer}{background}
            \node [fit=(noise_input), label={above:\textcolor{magenta}{\bfseries\textsc{Input Embed}}}] {}; 
        \end{pgfonlayer}

        \begin{pgfonlayer}{background}
            \node [inputemb, fit=(pointembed_input)(mplinear_input),] {}; 
        \end{pgfonlayer}

        \node[right=2.0cm of noise_input] (emb_block) {LevelEmb};

        \node[right=0.7cm of emb_block] (input_block) {InputEmb$_i$};

        \node[normalize, below=0.5cm of input_block] (normalize_block) {Normalize};
        \node[mpsilu, below=0.0cm of normalize_block] (silu1_block) {MPSiLU};
        \node[mplinear, below=0.0cm of silu1_block] (mplinear1_block) {MPLinear};

        \node[inner sep=0pt, below=0.3cm of mplinear1_block] (times_block) {$\bigotimes$};

        \node[mpsilu, below=0.3cm of times_block] (silu2_block) {MPSiLU};
        \node[mplinear, below=0.0cm of silu2_block] (mplinear2_block) {MPLinear};
        \node[mpadd, below=0.0cm of mplinear2_block] (mpadd_block) {MPAdd};

        \node[mplinear] at (emb_block |- mplinear1_block) (mplinear3_block) {MPLinear};

        \draw[arrow] (input_block) -- (normalize_block);
        \draw[arrow] (mplinear1_block) -- (times_block);
        \draw[arrow] (times_block) -- (silu2_block);

        \draw[arrow] (emb_block) -- (mplinear3_block);
        \coordinate[] (below_emb_block) at (times_block -| emb_block);
        \draw[thick] (mplinear3_block) -- (below_emb_block);

        \node[draw, inner sep=0pt, circle, left=0.7cm of times_block] (plus1_block) {+1};
        \draw[arrow] (below_emb_block) -- (plus1_block);
        \draw[arrow] (plus1_block) -- (times_block);

        \node[below=0.5cm of mpadd_block] (below_mpadd_block) {InputEmb$_{i+1}$};
        \draw[arrow] (mpadd_block) -- (below_mpadd_block);

        \coordinate[below=0.25cm of input_block] (below_input_block);
        \coordinate[right=0.5cm of input_block] (right_input_block);
        \coordinate[] (below_right_input_block) at (below_input_block -| right_input_block) ;
        \draw[arrow] (input_block) -- (below_input_block) -- (below_right_input_block) |- (mpadd_block);

        \coordinate[right=0.7cm of mpadd_block] (right_mpadd_block);

        \begin{pgfonlayer}{background}
            \node [block, fit=(below_right_input_block)(mplinear3_block)(normalize_block)(mpadd_block)(right_mpadd_block)] {}; 
        \end{pgfonlayer}

        \begin{pgfonlayer}{background}
            \node [fit=(emb_block)(input_block)(below_right_input_block), label={above:\textcolor{textblue}{\bfseries\textsc{Middle Block}}}] {}; 
        \end{pgfonlayer}

        \coordinate[right=0.8cm of mesh_denoiser, yshift=0.2cm] (right_mesh_denoiser) {};
        \coordinate[] (below_right_mesh_denoiser) at (right_mesh_denoiser |- output_denoiser);
        \draw[thick] (right_mesh_denoiser) -- (below_right_mesh_denoiser);

        \node[right=2.0cm of input_block] (emb_final) {LevelEmb};

        \node[right=0.7cm of emb_final] (input_final) {InputEmb$_d$};

        \node[normalize, below=0.5cm of input_final] (normalize_final) {Normalize};
        \node[mpsilu, below=0.0cm of normalize_final] (silu1_final) {MPSiLU};
        \node[mplinear, below=0.0cm of silu1_final] (mplinear1_final) {MPLinear};

        \node[inner sep=0pt, below=0.3cm of mplinear1_final] (times_final) {$\bigotimes$};

        \node[mpsilu, below=0.3cm of times_final] (silu2_final) {MPSiLU};
        \node[mplinear, below=0.0cm of silu2_final] (mplinear2_final) {MPLinear};

        \node[mplinear] at (emb_final |- mplinear1_final) (mplinear3_final) {MPLinear};

        \draw[arrow] (input_final) -- (normalize_final);
        \draw[arrow] (mplinear1_final) -- (times_final);
        \draw[arrow] (times_final) -- (silu2_final);

        \draw[arrow] (emb_final) -- (mplinear3_final);
        \coordinate[] (below_emb_final) at (times_final -| emb_final);
        \draw[thick] (mplinear3_final) -- (below_emb_final);

        \node[draw, inner sep=0pt, circle, left=0.7cm of times_final] (plus1_final) {+1};
        \draw[arrow] (below_emb_final) -- (plus1_final);
        \draw[arrow] (plus1_final) -- (times_final);

        \node[below=0.5cm of mplinear2_final] (output_final) {Output};
        \draw[arrow] (mplinear2_final) -- (output_final);

        \coordinate[below=0.25cm of input_final] (below_input_final);
        \coordinate[right=0.5cm of input_final] (right_input_final);


        \begin{pgfonlayer}{background}
            \node [final, fit=(mplinear3_final)(normalize_final)(mplinear2_final)] {}; 
        \end{pgfonlayer}

        \begin{pgfonlayer}{background}
            \node [fit=(emb_final)(input_final), label={above:\textcolor{mypurple}{\bfseries\textsc{Final Block}}}] {}; 
        \end{pgfonlayer}
    \end{tikzpicture}
}

%% file: tables/tab_network_arch.tex


\centering
\footnotesize
{\def\arraystretch{1.42}\tabcolsep=1.1em
\begin{tabular}{cc}
\toprule[1pt]
    \begin{tabular}[c]{@{}c@{}}\mytabtext{network}\\ \mytabtext{arch}\end{tabular}  &  
    \begin{tabular}[c]{@{}c@{}}\mytabtext{Chamfer $\downarrow$}\\ \mytabtext{ ($\times 10^3$)}\end{tabular}  \\ \midrule[0.5pt]
    Hashing Grid & 4.133 \\
    \rowcolor{tabcol!30}MLP & 2.647 \\ 
    Proposed & \textbf{2.140} \\ 
    \bottomrule[1pt]
\end{tabular}
}


%% file: tables/tab_ablation_data_size.tex
\centering \footnotesize
{\def\arraystretch{1}\tabcolsep=1.8em 
\begin{tabular}{cc}
\toprule[1pt]
\begin{tabular}[c]
{@{}c@{}}\mytabtext{dataset}\\ \mytabtext{size}\end{tabular}  &  
\begin{tabular}[c]{@{}c@{}}\mytabtext{Chamfer $\downarrow$}\\ \mytabtext{ ($\times 10^3$)}\end{tabular}  \\ \midrule[0.5pt]
    $10^5$ & 4.696 \\
    \rowcolor{tabcol!30}$10^6$ & 3.004 \\ 
    $10^7$ & 2.936 \\ 
    \rowcolor{tabcol!30}$10^8$ & 2.917 \\ 
    $\approx\infty$ & \textbf{2.916} \\ 
    \bottomrule[1pt]
\end{tabular}
}

%% file: tables/tab_sampling_steps.tex
\centering
\footnotesize
{\def\arraystretch{1.15}\tabcolsep=1.25em
\begin{tabular}{cc}
\toprule[1pt]
    \begin{tabular}[c]{@{}c@{}}\mytabtext{\# sampling}\\ \mytabtext{steps}\end{tabular} &
    \begin{tabular}[c]{@{}c@{}}\mytabtext{Chamfer $\downarrow$}\\ \mytabtext{ ($\times 10^3$)}\end{tabular}  \\ \midrule[0.5pt]
    8 & 9.803 \\
    \rowcolor{tabcol!30}16 & 2.814 \\ 
    32 & 2.782 \\ 
    \rowcolor{tabcol!30}64 & \textbf{2.780}\\
    \bottomrule[1pt]
\end{tabular}
}

%% file: tables/tab_ablation_block_size.tex
\centering \footnotesize
{\def\arraystretch{1}\tabcolsep=1.4em 
\begin{tabular}{cc}
\toprule[1pt]
    \begin{tabular}[c]{@{}c@{}}\mytabtext{\# network}\\ \mytabtext{blocks}\end{tabular}  &  
    \begin{tabular}[c]{@{}c@{}}\mytabtext{Chamfer $\downarrow$}\\ \mytabtext{ ($\times 10^3$)}\end{tabular}  \\ \midrule[0.5pt]
    2 & 3.784 \\
    \rowcolor{tabcol!30}4 & 3.565 \\ 
    6 & 3.547 \\
    \rowcolor{tabcol!30}8 & 3.546 \\ 
    10 & \textbf{3.544}\\
    \bottomrule[1pt]
\end{tabular}

}

%% file: tables/tab_gaussian_uniform.tex
\centering
\footnotesize
{\def\arraystretch{1.2}\tabcolsep=1.4em 
    \begin{tabular}{ccccc}
    \toprule[1pt]
       object & Wukong & lamp & lion & Parthenon \\ \midrule
       \mytabtext{Uniform} & 2.845 & 3.574& 3.734 & 7.621\\
       \rowcolor{tabcol!30} \mytabtext{Gaussian}  & \textbf{2.715} & \textbf{3.538} & \textbf{3.246} & \textbf{7.295} \\
       \bottomrule[1pt]
    \end{tabular}
}

%% file: tables/tab_inversion_steps.tex
\centering\footnotesize
{\def\arraystretch{1.2}\tabcolsep=0.35em 
    \begin{tabular}{cccccc}
    \toprule[1pt]
       \mytabtext{\# Inversion Steps}  & 4 & 8 & 16 & 32 & 64 \\ \midrule[0.5pt]
       \rowcolor{tabcol!30} 
       \mytabtext{MSE} $\downarrow$ & $6.88\mathrm{e}{-1}$ & $8.52\mathrm{e}{-2}$ & $6.14\mathrm{e}{-3}$ & $1.84\mathrm{e}{-4}$ & \textbf{$\mathbf{1.76\mathrm{e}{-6}}$} \\
       \bottomrule[1pt]
    \end{tabular}
}


%% file: bib.bib
@String(CVPR= {IEEE Conf. Comput. Vis. Pattern Recog.})

@String(TOG= {ACM Trans. Graph.})

@String(CVPR  = {CVPR})

@String(TOG   = {ACM TOG})

@article{liu2022flow,
  title={Flow straight and fast: Learning to generate and transfer data with rectified flow},
  author={Liu, Xingchao and Gong, Chengyue and Liu, Qiang},
  journal={arXiv preprint arXiv:2209.03003},
  year={2022}
}

@article{karras2022elucidating,
  title={Elucidating the design space of diffusion-based generative models},
  author={Karras, Tero and Aittala, Miika and Aila, Timo and Laine, Samuli},
  journal={Advances in neural information processing systems},
  volume={35},
  pages={26565--26577},
  year={2022}
}

@article{song2020denoising,
  title={Denoising diffusion implicit models},
  author={Song, Jiaming and Meng, Chenlin and Ermon, Stefano},
  journal={arXiv preprint arXiv:2010.02502},
  year={2020}
}

@inproceedings{Karras2024edm2,
  title     = {Analyzing and Improving the Training Dynamics of Diffusion Models},
  author    = {Tero Karras and Miika Aittala and Jaakko Lehtinen and
               Janne Hellsten and Timo Aila and Samuli Laine},
  booktitle = {Proc. CVPR},
  year      = {2024},
}

@article{zhang20223dilg,
  title={3dilg: Irregular latent grids for 3d generative modeling},
  author={Zhang, Biao and Nie{\ss}ner, Matthias and Wonka, Peter},
  journal={Advances in Neural Information Processing Systems},
  volume={35},
  pages={21871--21885},
  year={2022}
}

@article{bernardini1999ball,
  title={The ball-pivoting algorithm for surface reconstruction},
  author={Bernardini, Fausto and Mittleman, Joshua and Rushmeier, Holly and Silva, Cl{\'a}udio and Taubin, Gabriel},
  journal={IEEE transactions on visualization and computer graphics},
  volume={5},
  number={4},
  pages={349--359},
  year={1999},
  publisher={IEEE}
}

@article{levoy1985use,
  title={The use of points as a display primitive},
  author={Levoy, Marc and Whitted, Turner},
  year={1985},
  publisher={Citeseer}
}

@inproceedings{pfister2000surfels,
  title={Surfels: Surface elements as rendering primitives},
  author={Pfister, Hanspeter and Zwicker, Matthias and Van Baar, Jeroen and Gross, Markus},
  booktitle={Proceedings of the 27th annual conference on Computer graphics and interactive techniques},
  pages={335--342},
  year={2000}
}

@inproceedings{zwicker2001surface,
  title={Surface splatting},
  author={Zwicker, Matthias and Pfister, Hanspeter and Van Baar, Jeroen and Gross, Markus},
  booktitle={Proceedings of the 28th annual conference on Computer graphics and interactive techniques},
  pages={371--378},
  year={2001}
}

@article{yifan2019differentiable,
  title={Differentiable surface splatting for point-based geometry processing},
  author={Yifan, Wang and Serena, Felice and Wu, Shihao and {\"O}ztireli, Cengiz and Sorkine-Hornung, Olga},
  journal={ACM Transactions on Graphics (TOG)},
  volume={38},
  number={6},
  pages={1--14},
  year={2019},
  publisher={ACM New York, NY, USA}
}

@article{kerbl20233d,
  title={3D Gaussian Splatting for Real-Time Radiance Field Rendering.},
  author={Kerbl, Bernhard and Kopanas, Georgios and Leimk{\"u}hler, Thomas and Drettakis, George},
  journal={ACM Trans. Graph.},
  volume={42},
  number={4},
  pages={139--1},
  year={2023}
}

@inproceedings{mescheder2019occupancy,
  title={Occupancy networks: Learning 3d reconstruction in function space},
  author={Mescheder, Lars and Oechsle, Michael and Niemeyer, Michael and Nowozin, Sebastian and Geiger, Andreas},
  booktitle={Proceedings of the IEEE/CVF conference on computer vision and pattern recognition},
  pages={4460--4470},
  year={2019}
}

@inproceedings{park2019deepsdf,
  title={Deepsdf: Learning continuous signed distance functions for shape representation},
  author={Park, Jeong Joon and Florence, Peter and Straub, Julian and Newcombe, Richard and Lovegrove, Steven},
  booktitle={Proceedings of the IEEE/CVF conference on computer vision and pattern recognition},
  pages={165--174},
  year={2019}
}

@article{vecset,
author = {Zhang, Biao and Tang, Jiapeng and Nie\ss{}ner, Matthias and Wonka, Peter},
title = {{3DShape2VecSet}: A 3D Shape Representation for Neural Fields and Generative Diffusion Models},
year = {2023},
issue_date = {August 2023},
publisher = {Association for Computing Machinery},
address = {New York, NY, USA},
volume = {42},
number = {4},
issn = {0730-0301},
url = {https://doi.org/10.1145/3592442},
doi = {10.1145/3592442},
journal = {ACM Trans. Graph.},
month = {jul},
articleno = {92},
numpages = {16}
}

@article{chibane2020neural,
  title={Neural unsigned distance fields for implicit function learning},
  author={Chibane, Julian and Pons-Moll, Gerard and others},
  journal={Advances in Neural Information Processing Systems},
  volume={33},
  pages={21638--21652},
  year={2020}
}

@article{ho2020denoising,
  title={Denoising diffusion probabilistic models},
  author={Ho, Jonathan and Jain, Ajay and Abbeel, Pieter},
  journal={Advances in neural information processing systems},
  volume={33},
  pages={6840--6851},
  year={2020}
}

@article{lipman2022flow,
  title={Flow matching for generative modeling},
  author={Lipman, Yaron and Chen, Ricky TQ and Ben-Hamu, Heli and Nickel, Maximilian and Le, Matt},
  journal={arXiv preprint arXiv:2210.02747},
  year={2022}
}

@article{chen2022neural,
  title={Neural dual contouring},
  author={Chen, Zhiqin and Tagliasacchi, Andrea and Funkhouser, Thomas and Zhang, Hao},
  journal={ACM Transactions on Graphics (TOG)},
  volume={41},
  number={4},
  pages={1--13},
  year={2022},
  publisher={ACM New York, NY, USA}
}

@article{zheng2023locally,
  title={Locally attentional sdf diffusion for controllable 3d shape generation},
  author={Zheng, Xin-Yang and Pan, Hao and Wang, Peng-Shuai and Tong, Xin and Liu, Yang and Shum, Heung-Yeung},
  journal={ACM Transactions on Graphics (ToG)},
  volume={42},
  number={4},
  pages={1--13},
  year={2023},
  publisher={ACM New York, NY, USA}
}

@article{xiong2024octfusion,
  title={OctFusion: Octree-based Diffusion Models for 3D Shape Generation},
  author={Xiong, Bojun and Wei, Si-Tong and Zheng, Xin-Yang and Cao, Yan-Pei and Lian, Zhouhui and Wang, Peng-Shuai},
  journal={arXiv preprint arXiv:2408.14732},
  year={2024}
}

@inproceedings{ren2024xcube,
  title={Xcube: Large-scale 3d generative modeling using sparse voxel hierarchies},
  author={Ren, Xuanchi and Huang, Jiahui and Zeng, Xiaohui and Museth, Ken and Fidler, Sanja and Williams, Francis},
  booktitle={Proceedings of the IEEE/CVF Conference on Computer Vision and Pattern Recognition},
  pages={4209--4219},
  year={2024}
}

@inproceedings{yariv2024mosaic,
  title={Mosaic-sdf for 3d generative models},
  author={Yariv, Lior and Puny, Omri and Gafni, Oran and Lipman, Yaron},
  booktitle={Proceedings of the IEEE/CVF Conference on Computer Vision and Pattern Recognition},
  pages={4630--4639},
  year={2024}
}

@inproceedings{hui2022neural,
  title={Neural wavelet-domain diffusion for 3d shape generation},
  author={Hui, Ka-Hei and Li, Ruihui and Hu, Jingyu and Fu, Chi-Wing},
  booktitle={SIGGRAPH Asia 2022 Conference Papers},
  pages={1--9},
  year={2022}
}

@inproceedings{shue20233d,
  title={3d neural field generation using triplane diffusion},
  author={Shue, J Ryan and Chan, Eric Ryan and Po, Ryan and Ankner, Zachary and Wu, Jiajun and Wetzstein, Gordon},
  booktitle={Proceedings of the IEEE/CVF Conference on Computer Vision and Pattern Recognition},
  pages={20875--20886},
  year={2023}
}

@inproceedings{dong2024gpld3d,
  title={GPLD3D: Latent Diffusion of 3D Shape Generative Models by Enforcing Geometric and Physical Priors},
  author={Dong, Yuan and Zuo, Qi and Gu, Xiaodong and Yuan, Weihao and Zhao, Zhengyi and Dong, Zilong and Bo, Liefeng and Huang, Qixing},
  booktitle={Proceedings of the IEEE/CVF Conference on Computer Vision and Pattern Recognition},
  pages={56--66},
  year={2024}
}

@inproceedings{petrov2024gem3d,
  title={GEM3D: GEnerative Medial Abstractions for 3D Shape Synthesis},
  author={Petrov, Dmitry and Goyal, Pradyumn and Thamizharasan, Vikas and Kim, Vladimir and Gadelha, Matheus and Averkiou, Melinos and Chaudhuri, Siddhartha and Kalogerakis, Evangelos},
  booktitle={ACM SIGGRAPH 2024 Conference Papers},
  pages={1--11},
  year={2024}
}

@inproceedings{zhang2024functional,
  title={Functional diffusion},
  author={Zhang, Biao and Wonka, Peter},
  booktitle={Proceedings of the IEEE/CVF Conference on Computer Vision and Pattern Recognition},
  pages={4723--4732},
  year={2024}
}

@article{zeng2022lion,
  title={LION: Latent Point Diffusion Models for 3D Shape Generation},
  author={Zeng, Xiaohui and Vahdat, Arash and Williams, Francis and Gojcic, Zan and Litany, Or and Fidler, Sanja and Kreis, Karsten},
  journal={arXiv preprint arXiv:2210.06978},
  year={2022}
}

@inproceedings{luo2021diffusion,
  title={Diffusion probabilistic models for 3d point cloud generation},
  author={Luo, Shitong and Hu, Wei},
  booktitle={Proceedings of the IEEE/CVF conference on computer vision and pattern recognition},
  pages={2837--2845},
  year={2021}
}

@inproceedings{zhou20213d,
  title={3d shape generation and completion through point-voxel diffusion},
  author={Zhou, Linqi and Du, Yilun and Wu, Jiajun},
  booktitle={Proceedings of the IEEE/CVF international conference on computer vision},
  pages={5826--5835},
  year={2021}
}

@inproceedings{takikawa2021neural,
  title={Neural geometric level of detail: Real-time rendering with implicit 3d shapes},
  author={Takikawa, Towaki and Litalien, Joey and Yin, Kangxue and Kreis, Karsten and Loop, Charles and Nowrouzezahrai, Derek and Jacobson, Alec and McGuire, Morgan and Fidler, Sanja},
  booktitle={Proceedings of the IEEE/CVF Conference on Computer Vision and Pattern Recognition},
  pages={11358--11367},
  year={2021}
}

@article{martel2021acorn,
  title={Acorn: Adaptive coordinate networks for neural scene representation},
  author={Martel, Julien NP and Lindell, David B and Lin, Connor Z and Chan, Eric R and Monteiro, Marco and Wetzstein, Gordon},
  journal={arXiv preprint arXiv:2105.02788},
  year={2021}
}

@article{muller2022instant,
  title={Instant neural graphics primitives with a multiresolution hash encoding},
  author={M{\"u}ller, Thomas and Evans, Alex and Schied, Christoph and Keller, Alexander},
  journal={ACM transactions on graphics (TOG)},
  volume={41},
  number={4},
  pages={1--15},
  year={2022},
  publisher={ACM New York, NY, USA}
}

@article{sitzmann2020implicit,
  title={Implicit neural representations with periodic activation functions},
  author={Sitzmann, Vincent and Martel, Julien and Bergman, Alexander and Lindell, David and Wetzstein, Gordon},
  journal={Advances in neural information processing systems},
  volume={33},
  pages={7462--7473},
  year={2020}
}

@inproceedings{yang2023neural,
  title={Neural vector fields: Implicit representation by explicit learning},
  author={Yang, Xianghui and Lin, Guosheng and Chen, Zhenghao and Zhou, Luping},
  booktitle={Proceedings of the IEEE/CVF Conference on Computer Vision and Pattern Recognition},
  pages={16727--16738},
  year={2023}
}

@article{zhou2024cap,
  title={CAP-UDF: Learning Unsigned Distance Functions Progressively from Raw Point Clouds with Consistency-Aware Field Optimization},
  author={Zhou, Junsheng and Ma, Baorui and Li, Shujuan and Liu, Yu-Shen and Fang, Yi and Han, Zhizhong},
  journal={IEEE Transactions on Pattern Analysis and Machine Intelligence},
  year={2024},
  publisher={IEEE}
}

@inproceedings {meshlab, booktitle = {Eurographics Italian Chapter Conference}, editor = {Vittorio Scarano and Rosario De Chiara and Ugo Erra}, title = {{MeshLab: an Open-Source Mesh Processing Tool}}, author = {Cignoni, Paolo and Callieri, Marco and Corsini, Massimiliano and Dellepiane, Matteo and Ganovelli, Fabio and Ranzuglia, Guido}, year = {2008}, publisher = {The Eurographics Association}, ISBN = {978-3-905673-68-5}, DOI = {10.2312/LocalChapterEvents/ItalChap/ItalianChapConf2008/129-136} }

@inproceedings{guillard2022meshudf,
  title={Meshudf: Fast and differentiable meshing of unsigned distance field networks},
  author={Guillard, Benoit and Stella, Federico and Fua, Pascal},
  booktitle={European Conference on Computer Vision},
  pages={576--592},
  year={2022},
  organization={Springer}
}

@inproceedings{edavamadathil2024neural,
  title={Neural Geometry Fields For Meshes},
  author={Edavamadathil Sivaram, Venkataram and Li, Tzu-Mao and Ramamoorthi, Ravi},
  booktitle={ACM SIGGRAPH 2024 Conference Papers},
  pages={1--11},
  year={2024}
}

@article{liu2020neural,
  title={Neural sparse voxel fields},
  author={Liu, Lingjie and Gu, Jiatao and Zaw Lin, Kyaw and Chua, Tat-Seng and Theobalt, Christian},
  journal={Advances in Neural Information Processing Systems},
  volume={33},
  pages={15651--15663},
  year={2020}
}

@inproceedings{sun2022direct,
  title={Direct voxel grid optimization: Super-fast convergence for radiance fields reconstruction},
  author={Sun, Cheng and Sun, Min and Chen, Hwann-Tzong},
  booktitle={Proceedings of the IEEE/CVF conference on computer vision and pattern recognition},
  pages={5459--5469},
  year={2022}
}

@article{xiao2023unsupervised,
  title={Unsupervised point cloud representation learning with deep neural networks: A survey},
  author={Xiao, Aoran and Huang, Jiaxing and Guan, Dayan and Zhang, Xiaoqin and Lu, Shijian and Shao, Ling},
  journal={IEEE Transactions on Pattern Analysis and Machine Intelligence},
  volume={45},
  number={9},
  pages={11321--11339},
  year={2023},
  publisher={IEEE}
}

@article{guo2020deep,
  title={Deep learning for 3d point clouds: A survey},
  author={Guo, Yulan and Wang, Hanyun and Hu, Qingyong and Liu, Hao and Liu, Li and Bennamoun, Mohammed},
  journal={IEEE transactions on pattern analysis and machine intelligence},
  volume={43},
  number={12},
  pages={4338--4364},
  year={2020},
  publisher={IEEE}
}

@article{bello2020deep,
  title={Deep learning on 3D point clouds},
  author={Bello, Saifullahi Aminu and Yu, Shangshu and Wang, Cheng and Adam, Jibril Muhmmad and Li, Jonathan},
  journal={Remote Sensing},
  volume={12},
  number={11},
  pages={1729},
  year={2020},
  publisher={MDPI}
}

@article{botsch2010polygon,
  title={Polygon mesh processing},
  author={Botsch, Mario},
  journal={AK Peters},
  year={2010}
}

@article{bronstein2017geometric,
  author={Bronstein, Michael M. and Bruna, Joan and LeCun, Yann and Szlam, Arthur and Vandergheynst, Pierre},
  journal={IEEE Signal Processing Magazine}, 
  title={Geometric Deep Learning: Going beyond Euclidean data}, 
  year={2017},
  volume={34},
  number={4},
  pages={18-42},
  doi={10.1109/MSP.2017.2693418}}

@inproceedings{fey2018splinecnn,
  title={Splinecnn: Fast geometric deep learning with continuous b-spline kernels},
  author={Fey, Matthias and Lenssen, Jan Eric and Weichert, Frank and M{\"u}ller, Heinrich},
  booktitle={Proceedings of the IEEE conference on computer vision and pattern recognition},
  pages={869--877},
  year={2018}
}

@article{zhang2024gaussiancube,
  title={GaussianCube: Structuring Gaussian Splatting using Optimal Transport for 3D Generative Modeling},
  author={Zhang, Bowen and Cheng, Yiji and Yang, Jiaolong and Wang, Chunyu and Zhao, Feng and Tang, Yansong and Chen, Dong and Guo, Baining},
  journal={arXiv preprint arXiv:2403.19655},
  year={2024}
}

@article{roessle2024l3dg,
  title={L3DG: Latent 3D Gaussian Diffusion},
  author={Roessle, Barbara and M{\"u}ller, Norman and Porzi, Lorenzo and Bul{\`o}, Samuel Rota and Kontschieder, Peter and Dai, Angela and Nie{\ss}ner, Matthias},
  journal={arXiv preprint arXiv:2410.13530},
  year={2024}
}

@article{yifan2021geometry,
  title={Geometry-consistent neural shape representation with implicit displacement fields},
  author={Yifan, Wang and Rahmann, Lukas and Sorkine-Hornung, Olga},
  journal={arXiv preprint arXiv:2106.05187},
  year={2021}
}

@inproceedings{gu2002geometry,
  title={Geometry images},
  author={Gu, Xianfeng and Gortler, Steven J and Hoppe, Hugues},
  booktitle={Proceedings of the 29th annual conference on Computer graphics and interactive techniques},
  pages={355--361},
  year={2002}
}

@article{aigerman2022neural,
  title={Neural jacobian fields: Learning intrinsic mappings of arbitrary meshes},
  author={Aigerman, Noam and Gupta, Kunal and Kim, Vladimir G and Chaudhuri, Siddhartha and Saito, Jun and Groueix, Thibault},
  journal={arXiv preprint arXiv:2205.02904},
  year={2022}
}

@article{korosteleva2021generating,
  title={Generating datasets of 3d garments with sewing patterns},
  author={Korosteleva, Maria and Lee, Sung-Hee},
  journal={arXiv preprint arXiv:2109.05633},
  year={2021}
}

@article{zhou2016thingi10k,
  title={Thingi10k: A dataset of 10,000 3d-printing models},
  author={Zhou, Qingnan and Jacobson, Alec},
  journal={arXiv preprint arXiv:1605.04797},
  year={2016}
}

@inproceedings{yang2019pointflow,
  title={Pointflow: 3d point cloud generation with continuous normalizing flows},
  author={Yang, Guandao and Huang, Xun and Hao, Zekun and Liu, Ming-Yu and Belongie, Serge and Hariharan, Bharath},
  booktitle={Proceedings of the IEEE/CVF international conference on computer vision},
  pages={4541--4550},
  year={2019}
}

@inproceedings{cai2020learning,
  title={Learning gradient fields for shape generation},
  author={Cai, Ruojin and Yang, Guandao and Averbuch-Elor, Hadar and Hao, Zekun and Belongie, Serge and Snavely, Noah and Hariharan, Bharath},
  booktitle={Computer Vision--ECCV 2020: 16th European Conference, Glasgow, UK, August 23--28, 2020, Proceedings, Part III 16},
  pages={364--381},
  year={2020},
  organization={Springer}
}

@inproceedings{rezende2015variational,
  title={Variational inference with normalizing flows},
  author={Rezende, Danilo and Mohamed, Shakir},
  booktitle={International conference on machine learning},
  pages={1530--1538},
  year={2015},
  organization={PMLR}
}

@article{song2019generative,
  title={Generative modeling by estimating gradients of the data distribution},
  author={Song, Yang and Ermon, Stefano},
  journal={Advances in neural information processing systems},
  volume={32},
  year={2019}
}

@inproceedings{yushigaussiananything,
  title={GaussianAnything: Interactive Point Cloud Flow Matching for 3D Generation},
  author={Yushi, LAN and Zhou, Shangchen and Lyu, Zhaoyang and Hong, Fangzhou and Yang, Shuai and Dai, Bo and Pan, Xingang and Loy, Chen Change},
  booktitle={The Thirteenth International Conference on Learning Representations}
}

@article{davies2020effectiveness,
  title={On the effectiveness of weight-encoded neural implicit 3d shapes},
  author={Davies, Thomas and Nowrouzezahrai, Derek and Jacobson, Alec},
  journal={arXiv preprint arXiv:2009.09808},
  year={2020}
}

@article{guan2023learning,
  title={Learning neural implicit representations with surface signal parameterizations},
  author={Guan, Yanran and Chubarau, Andrei and Rao, Ruby and Nowrouzezahrai, Derek},
  journal={Computers \& Graphics},
  volume={114},
  pages={257--264},
  year={2023},
  publisher={Elsevier}
}
